\documentclass[twoside,11pt]{article}

\usepackage{jmlr2e}

\usepackage{graphicx} 
\usepackage{subfigure} 

\usepackage{natbib}

\usepackage{algorithm} 
\usepackage{algorithmic} 
\usepackage{amsmath}
\usepackage{amsfonts}
\usepackage{multirow}
\usepackage{array}
\usepackage{rotating}

\jmlrheading{1}{2000}{1-48}{4/00}{10/00}{Dongwoo Kim and Alice Oh}

\ShortHeadings{Hierarchical Dirichlet Scaling Process}{Kim and Oh}
\firstpageno{1}

\begin{document}

\title{Hierarchical Dirichlet Scaling Process}

\author{\name Dongwoo Kim\thanks{Corresponding author; email: dw.kim@kaist.ac.kr} \email dw.kim@kaist.ac.kr \\
       \addr Department of Computer Science\\
       KAIST\\
       Daejeon, Republic of Korea 
       \AND
       \name Alice Oh \email alice.oh@kaist.edu \\
       \addr Department of Computer Science\\
       KAIST\\
       Daejeon, Republic of Korea
       }

\editor{Unknown} 

\maketitle

\begin{abstract}
We present the \textit{hierarchical Dirichlet scaling process} (HDSP), a Bayesian nonparametric mixed membership model.
The HDSP generalizes the hierarchical Dirichlet process (HDP) to model the correlation structure between metadata in the corpus and mixture components.
We construct the HDSP based on the normalized gamma representation of the Dirichlet process, and this construction allows 
incorporating a scaling function that controls the membership probabilities of the mixture components.
We develop two scaling methods to demonstrate that different modeling assumptions can be expressed in the HDSP. 
We also derive the corresponding approximate posterior inference algorithms using variational Bayes. 
Through experiments on datasets of newswire, medical journal articles, conference proceedings, and product reviews, we show that the HDSP results in a better predictive performance than labeled LDA, partially labeled LDA, and author topic model and a better negative review classification performance than the supervised topic model and SVM.
\end{abstract}


\section{Introduction}

The hierarchical Dirichlet process (HDP) is an important nonparametric Bayesian prior for mixed membership models, and the HDP topic model is useful for a wide variety of tasks involving unstructured text \citep{Teh:2006p3792}. To extend the HDP topic model, there has been active research in dependent random probability measures as priors for modeling the underlying association between the latent semantic structure and covariates, such as time stamps and spatial coordinates \citep{ahmed2012timeline, ren2011logistic}. 

A large body of this research is rooted in the dependent Dirichlet process (DDP) \citep{maceachern1999dependent} where the probabilistic random measure is defined as a function of covariates. Most DDP approaches rely on the generalization of Sethuraman's stick breaking representation of DP \citep{sethuraman1991constructive}, incorporating the time difference between two or more data points, the spatial difference among observed data, or the ordering of the data points into the predictor dependent stick breaking process \citep{duan2007generalized,dunson2008kernel, RePEc:bes:jnlasa:v:101:y:2006:p:179-194}. Some of these priors can be integrated into the hierarchical construction of DP \citep{srebro2005time}, resulting in topic models where temporally- or spatially-proximate data are more likely to be clustered.

These existing DP approaches, however, cannot model datasets with various types of covariates, including categorical and numerical labels. One reason is that categorical labels cannot be used to directly define the similarity between two documents, unlike temporal or spatial information. Also, labels and documents do not have a one-to-one correspondence, as there may be zero, one, or more labels per document. Furthermore, existing DP approaches cannot be applied to datasets with more than one type of covariates, for example numerical and categorical labels. 

We suggest the \textit{hierarchical Dirichlet scaling process} (HDSP) as a new way of modeling a corpus with various types of covariates such as categories, authors, and numerical ratings. The HDSP models the relationship between topics and covariates by generating dependent random measures in a hierarchy, where the first level is a Dirichlet process, and the second level is a \textit{Dirichlet scaling process} (DSP). The first level DP is constructed in the traditional way of a stick breaking process, and the second level DSP with a normalized gamma process.
With the normalized gamma process, each topic proportion of a document is independently drawn from a gamma distribution and then normalized. Unlike the stick breaking process, the normalized gamma process keeps the same order of the atoms as the first level measure, which allows the topic proportions in the random measure to be controlled.
The DSP then uses that controllability to guide the topic proportions of a document by replacing the rate parameter of the gamma distribution with a scaling function that defines the correlation structure between topics and labels. 
The choice of the scaling function reflects the characteristics of the corpus. 
We show two scaling functions, the first one for a corpus with categorical labels, and the second for a corpus with both categorical and numerical labels.



The HDSP models the topic proportions of a document as a dependent variable of observable side information. This modeling approach differs from the traditional definition of a generative process where the observable variables are generated from a latent variable or parameter. For example, \cite{zhu2009medlda} and  \cite{blei2010supervised} propose generative processes where the observable labels are generated from a topic proportion of a document. However, a more natural model of the human writing process is to decide what to write about (e.g., categories) before writing the content of a document. This same approach is also successfully demonstrated in \cite{mimno2012topic}.

The outline of this paper is as follows. In Section \ref{related}, we describe related work and position our work within the topic modeling literature. In Section \ref{sec:hdsp}, we describe the gamma process construction of the HDP and how scale parameters are used to develop the HDSP with two different scaling functions. In Section \ref{sec:vi}, we derive a variational inference for the latent variables. In Section \ref{sec:exp}, we verify our approach on a synthetic dataset and demonstrate the improved predictive power on real world corpora. In Section \ref{sec:con}, we discuss our conclusions and possible directions for future work.

\section{\label{related}Related Work}
For model construction, the model most closely related to HDSP is the discrete infinite logistic normal (DILN) model \citep{paisley2012discrete} in which the correlations among topics are modeled through the normalized gamma construction. DILN allocates a latent location for each topic in the first level, and then draws the second level random measures from the normalized  gamma construction of the DP. Those random measures are then scaled by an exponentiated Gaussian process defined on the latent locations. DILN is a nonparametric counterpart of the correlated topic model \citep{blei2007correlated} in which the logistic normal prior is used to model the correlations between topics. The HDSP is also constructed through the normalized gamma distribution with an informative scaling parameter, but our goal in HDSP is to model the correlations between topics and labels.

The Dirichlet-multinomial regression topic model (DMR-TM) \citep{mimno2012topic} also models the label dependent topic proportions of documents, but it is a parametric model. The DMR-TM places a log-linear prior on the parameter of the Dirichlet distribution to incorporate arbitrary types of observed labels. The DMR-TM takes the ``upstream'' approach in which the latent variable or latent topics are conditionally generated from the observed label information. The author-topic model \citep{Rosen-Zvi:2004:AMA:1036843.1036902} also takes the same approach, but it is a specialized model for authors of documents. Unlike the  ``downstream'' generative approach used in the supervised topic model \citep{blei2010supervised}, the maximum margin topic model \citep{zhu2009medlda}, and the relational topic model \citep{chang2009relational}, the upstream approach does not require specifying the probability distribution over all possible values of observed labels.

The HDSP is a new way of constructing a dependent random measure in a hierarchy. In the field of Bayesian nonparametrics, the introduction of DDP \citep{sethuraman1991constructive} has led to increased attention in constructing dependent random measures. Most such approaches develop priors to allow covariate dependent variation in the atoms of the random measure \citep{gelfand2005bayesian, rao2009spatial} or in the weights of atoms \citep{RePEc:bes:jnlasa:v:101:y:2006:p:179-194, duan2007generalized, dunson2008kernel}. These priors replace the first level of the HDP to incorporate a document-specific covariate for generating a dependent topic proportion. These approaches focus on the spatial distances or the ordering of the covariate, so they cannot be generalized for arbitrary types of label information. The HDSP, on the other hand, can model any types of labels. The HDSP allows covariate dependent variation in the weights of atoms, where the variation is controlled by the scaling function that defines the correlation between atoms and labels. A proper definition of the scaling function gives the flexibility to model various types of labels.

Several topic models for labeled documents use the credit attribution approach where each observed word token is assigned to one of the observed labels. Labeled LDA (L-LDA) allocates one dimension of the topic simplex per label and generates words from only the topics that correspond to the labels in each document \citep{ramage2009labeled}. An extension of this model, partially labeled LDA (PLDA), adds more flexibility by allocating a pre-defined number of topics per label and including a background label to handle documents with no labels \citep{Ramage:2011:PLT:2020408.2020481}. The Dirichlet process with mixed random measures (DP-MRM) is a nonparametric topic model which generates an unbounded number of topics per label but still excludes topics from labels that are not observed in the document \citep{kim2012icml}. 

\section{\label{sec:hdsp}Hierarchical Dirichlet Scaling Process}
In this section, we describe the hierarchical Dirichlet scaling process (HDSP). First we review the HDP with an alternative construction using the normalized gamma process construction for the second level DP. We then present the HDSP where the second level DP is replaced by Dirichlet scaling process (DSP). Finally, we describe two scaling functions for the DSP to incorporate categorical and numerical labels.

\subsection{The normalized gamma process construction of HDP}
The HDP\footnote{In this paper, we limit our discussions of the HDP to the two level construction of the DP and refer to it simply as the HDP.} consists of  two levels of the DP where the random measure drawn from the upper level DP is the base distribution of the lower level DP. The formal definition of the hierarchical representation is as follows:
\begin{align}
G_0 \sim \text{DP}(\alpha, H), \qquad{} G_m \sim \text{DP}(\beta, G_0),
\end{align}
where $H$ is a base distribution, $\alpha$, and $\beta$ are concentration parameters for each level respectively, and index $m$ represents multiple draws from the second level DP. For the mixed membership model, $\mathrm{x}_{mn}$, observation $n$ in group $m$,  can be drawn from
\begin{align}
\theta_{mn} \sim G_m, \qquad{} \mathrm{x}_{mn} \sim f(\theta_{mn}),
\end{align}
where $f(\cdot)$ is a data distribution parameterized by $\theta$. In the context of topic models, the base distribution $H$ is usually a Dirichlet distribution over the vocabulary, so the atoms of the first level random measure $G_0$ are an infinite set of topics drawn from $H$. The second level random measure $G_m$ is distributed based on the first level random measure $G_0$, so the second level shares the same set of topics, the atoms of the first level random measure.

The constructive definition of the DP can be represented as a stick breaking process \citep{sethuraman1991constructive}, 
and in the HDP inference algorithm based on stick breaking, the first level DP is given by the following conditional distributions:
\begin{align}
&V_k \sim \text{Beta}(1, \alpha) & p_k = V_k \prod_{j=1}^{j<k} (1-V_j)&\notag\\
&\phi_k \sim H & G_0 = \sum_{k=1}^{\infty} p_k \delta_{\phi_k},&
\end{align}
where $V_k$ defines a corpus level topic distribution for topic $\phi_k$. The second level random measures are conditionally distributed on the first level discrete random measure $G_0$:
\begin{align}
&\pi_{ml} \sim \text{Beta}(1, \beta) & p_{ml} = \pi_{ml} \prod_{j=1}^{j<l} (1-\pi_{mj})& \label{eqn:second}\notag\\
&\theta_{ml} \sim G_0 & G_m = \sum_{l=1}^{\infty} p_{ml} \delta_{\theta_{ml}},&
\end{align}
where the second level atom $\theta_{ml}$ corresponds to one of the first level atoms $\phi_k$. 
This stick breaking construction is the most widely used method for the hierarchical construction \citep{wang2011online, Teh:2006p3792}. 

An alternative construction of the HDP is based on the normalized gamma process \citep{paisley2012discrete}. While the first level construction remains the same, the gamma process changes the second level construction from Eq. \ref{eqn:second} to
\begin{align}
\pi_{mk} &\sim \text{Gamma}(\beta p_k, 1)  \notag \\
G_m &= \sum_{k=1}^{\infty} \frac{ \pi_{mk} }{\sum_{j=1}^{\infty} \pi_{mj}} \delta \phi_k,
\end{align}
where Gamma$(x;a, b) = b^a x^{(a-1)} e^{-bx} / \Gamma(a)$. 
Unlike the stick breaking construction, the atom of the $\pi_{mk}$ of the gamma process is the same as the atom of the $k$th stick of the first level. Therefore, during inference, the model does not need to keep track of which second level atoms correspond to which first level atoms. Furthermore, by placing a proper random variable on the rate parameter of the gamma distribution, the model can infer the correlations among the topics \citep{paisley2012discrete} through the Gaussian process \citep{Rasmussen:2005:GPM:1162254}. 

The normalized gamma process itself is not an appropriate construction method for the approximate posterior inference algorithm based on the variational truncation method \citep{blei2006variational} because, unlike the stick breaking process, the probability mass of a random measure constructed by the normalized gamma process is not limited to the first few number of atoms. But once the base distribution of second level DP is constructed by the stick breaking process of first level DP, the total mass of the second level base distribution $G_0$ is limited to the first few number of atoms, and then the truncation based posterior inference algorithm approximates the true posterior of the normalized gamma construction. 

\begin{figure*}[t!]
	\centering
	\includegraphics[width=\linewidth, trim=0 0 0 0, clip=true]{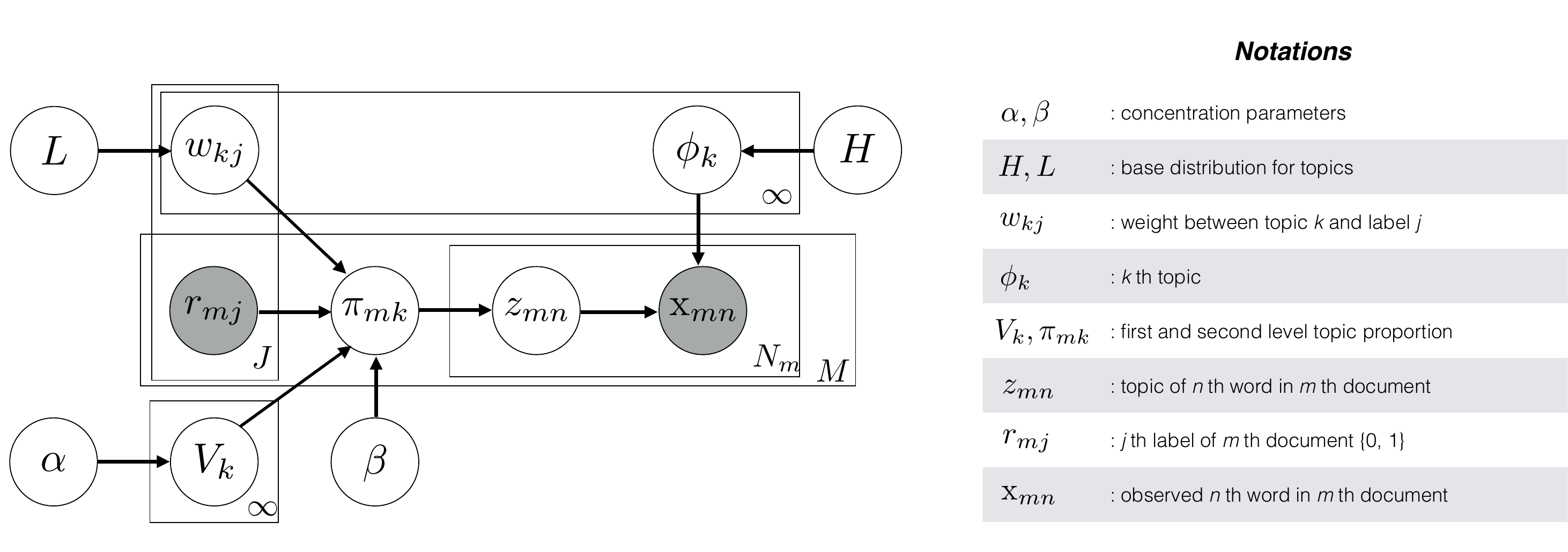}		
	\caption{\label{fig:graphmodel}Graphical model of the hierarchical Dirichlet scaling process.}
\end{figure*}

\subsection{Hierarchical Dirichlet scaling process}
The HDSP generalizes the HDP by modeling mixture proportions dependent on covariates.
As a topic model, the HDSP assumes that topics and labels are correlated, and the topic proportions of a document are proportional to the correlations between the topics and the observed labels of the document. 
We develop the Dirichlet scaling process (DSP) with the normalized gamma construction of the DP, where the rate parameter of the gamma distribution is replaced by the scaling function. This scaling function serves the central role of defining the correlation structure between a topic and labels. 
Formally, the HDSP consists of DP and DSP in a hierarchy:
\begin{align}
G_0 &\sim \text{DP}(\alpha, H) \\
G_m &\sim \text{DSP}(\beta, G_0, r_{m}, s_w(\cdot)),
\end{align}
where the first level random measure $G_0$ is drawn from the DP with concentration parameter $\alpha$ and base distribution $H$. The second level random measure $G_m$ for document $m$ is drawn from the DSP parameterized by the concentration parameter $\beta$, base distribution $G_0$, observed labels of document $r_{m}$, and \textit{scaling function} $s(\cdot)$ with \textit{scaling parameter} $w$. 

As in the HDP, the first level of HDSP is a DP where the base distribution is the product of two distributions for data distribution and scaling parameter $w$. Specifically, the base distribution $H$ is $\text{Dir}(\eta) \otimes L_w$ where $\eta$ is the parameter of the word-topic distribution, and $L_w$ is a prior distribution for the scaling parameter $w$. The form of the resulting random measure is
\begin{align}
G_0 = \sum_{k=1}^{\infty} p_k \delta_{\{\phi_k, w_k\}},
\end{align}
where $p_k$ is the stick length for topic $k$, $p_k = V_k \prod_{k'=1}^{k'<k}(1-V_{k'})$ and \{$\phi_k$, $w_k$\} is the atom of stick $k$. At the second level construction, $w_k$ becomes the parameter to guide the proportion of topic $k$'s for each document. 


At the second level of HDSP, label-dependent random measures are drawn from the DSP. First, as in the HDP, draw a random measure $G_m' \sim \text{DP}(\beta, H)$ for document $m$. Second, scale the weights of the atoms based on a scaling function parameterized by $w_k$ and the observed labels. Let $r_{mj}$ be the value of observed label $j$ in document $m$, then $G_m'$ is scaled as follows:
\begin{align}
G_m(\{\phi_k, l_k\}) \propto G_m'(\{\phi_k, l_k\}) \times s_{w_k}(r_{mj})
\end{align}
where $s_{w_k}(\cdot)$ is the scaling function parameterized by the scaling parameter $w_k$. Topic $k$ is scaled by the scaling weight, $s_{w_k}(r_{mj})$, and therefore, the topic proportions of a document is proportional to the scaling weights of the observed labels. The scaling function should be carefully chosen to reflect the underlying relationship between topics and labels. We show two concrete examples of scaling functions in Section \ref{sec:sfn}.


The constructive definition of HDSP is similar to the HDP, but the difference comes from the scaling function. The stick breaking process is used to construct the first level random measure:
\begin{align}
&V_k \sim \text{Beta}(1, \alpha) & p_k = V_k \prod_{j=1}^{j < k} (1-V_j)& \label{eqn:st1} \notag\\
&\phi_k \sim \text{Dir}(\eta),\quad w_k \sim L_w & G_0 = \sum_{k=1}^{\infty} p_k \delta_{\{\phi_k, w_k\}},&
\end{align}
where the pair \{$\phi_k$, $w_k$\} drawn i.i.d. from two base distributions forms an atom of the resulting measure.

Based on the discrete first level random measure, the second level random measure is constructed by the normalized gamma process. As in the HDP, the weight of atom $k$ is drawn from a gamma distribution with parameter $\beta p_k$, and then scaled by the scaling weight $s_{w_k}(r_m)$
\begin{align}
\pi_{mk} &\sim \text{Gamma}(\beta p_k, 1) \times s_{w_k}(r_m).
\end{align}
The scaling weight can be directly incorporated into the second parameter of the gamma distribution because the scaled gamma random variable $y = kx \sim$ Gamma$(a,1)$ is equal to $y \sim$ Gamma$(a, k^{-1})$,
\begin{align}
\pi_{mk} &\sim \text{Gamma}(\beta p_k, s_{w_k}(r_m)^{-1}). \notag\\
\end{align}
Then, the random variables are normalized to form a proper probability random measure
\begin{align}
G_m &= \sum_{k=1}^{\infty} \frac{\pi_{mk}}{\sum_{j=1}^{\infty} \pi_{mj}} \delta_{\phi_k}. \label{eqn:st4}
\end{align}
For the mixed membership model, $n$th observation in $m$th group is drawn as follows:
\begin{align}
\phi_k \sim G_m, \quad \mathrm{x}_{mn} \sim f(\phi_k), \label{eqn:mmt}
\end{align}
where $f$ is a data distribution parameterized by $\phi_k$.
For topic modeling, $G_m$ and $x_{mn}$ correspond to document $m$ and word $n$ in document $m$, respectively. 

\subsection{\label{sec:sfn}Scaling functions}

Now we propose two scaling functions to express the correlation between topics and labels of documents. A scaling method is properly defined by two factors: 1) a proper prior over the scaling parameter $w_k$, 2) a plausible scaling function between topic specific scaling parameter $w_k$ and the observed labels of document $r_m$. 

\textbf{Scaling function 1}: We design the first scaling function to model categorical side information such as authors, tags, and categories. For a corpus with $J$ unique labels, then $w_k$ is a $J$-dimensional parameter where each dimension matches to a corresponding label. We define the scaling function as the product of scaling parameters that correspond to the observed labels:
\begin{align}
\label{eqn:scaling_fn1}
s_{w_k}(r_m) = \prod_{j=1}^{J} w_{kj}^{r_{mj}} \quad \quad \quad w_{kj} \sim \text{inv-Gamma}(a_w, b_w)
\end{align}
where $r_{mj}$ is an indicator variable whose value is one when label $j$ is observed in document $m$ and zero otherwise. $w_{kj}$ is a scaling parameter of topic $k$ for label $j$. We place a inverse gamma prior over the weight variable $w_{kj}$. 

With this scaling function, the proportion of topic $k$ for document $m$ is scaled as follows:
\begin{align}
\pi_{mk} \sim \text{Gamma}(\beta p_k, 1) \times \prod_{j=1}^{J} w_{kj}^{r_{mj}}.
\end{align}
The scaled gamma distribution is equal to the gamma distribution with the rate parameter of inverse scaling factor, so we can rewrite the above equation as follows:
\begin{align}
\pi_{mk} &\sim \text{Gamma}(\beta p_k, \prod_{j=1}^{J} w_{kj}^{-r_{mj}}).
\end{align}

Finally, we normalize these random variables to make a probabilistic random measure summed up to unity for document $m$:
\begin{align}
\bar\pi_{mk} = \frac{\pi_{mk}}{\sum_{k'} \pi_{mk'}}.
\end{align}

\textbf{Scaling function 2}: The above scaling function models categorical side information, but many datasets, such as product reviews have numerical ratings as well as categorical information. We propose the second scaling function that can model both numerical and categorical information. Again, let $w_k$ be $J$-dimensional scaling parameter where each dimension matches to a corresponding label. The second scaling function is defined as follows:
\begin{align}
s_{w_k}(r_m) = \frac{1}{\exp(\sum_{j}w_{kj}r_{mj})},
\end{align}
where $w_{kj}$ is the scaling parameter of label $j$ for topic $k$, and $r_{mj}$ is the observed value of label $j$ of document $m$. We place a normal prior over the scaling parameter $w_{k}$. The scaling function is an inverse log-linear to the weighted sum of document's labels. Unlike the previous scaling function which only considers whether a label is observed in a document, this scaling function incorporates the value of the observed label. With this scaling function, the proportion of topic $k$ for document $m$ is scaled as follows
\begin{align}
\pi_{mk} \sim \text{Gamma}(\beta p_k, 1) \times \frac{1}{\exp(\sum_{j}w_{kj}r_{mj})}.
\end{align}
Again, we can rewrite this equations as
\begin{align}
\label{eqn:scaling_fn}
\pi_{mk} \sim \text{Gamma}(\beta p_k, \exp(\sum_{j}w_{kj}r_{mj})).
\end{align}
$\pi_{mk}$ is proportional to the inverse weighted sum of observed labels. Again, we normalize $\pi_{mk}$ to construct a proper random measure.

The choice of scaling function reflects the modeler's perspective with respect to the underlying relationship between topics and labels. 
The first scaling function scales each topic by the product of the scaling parameters of the observed labels.
This reflects the modeler's assumption that a document with a set of observed labels is likely to exhibit topics that have high correlation with all of the observed labels.
With the second scaling function, the scaling weight changes exponentially as the value of label changes. 
This reflects the modeler's assumption that two documents with the same set of observed labels but with different values are likely to exhibit different topics. 

\subsection{HDSP as a dependent Dirichlet process}

We can view the HDSP as an alternative construction of the hierarchical dependent Dirichlet process (DDP) via a hierarchy consisting of a stick breaking process and a normalized gamma process. Let us compare the HDSP approach to the general DDP approach for topic modeling. The formal definition of DDP is:
\begin{align}
G_0(\cdot) &\sim \text{DDP}(\alpha, H),
\end{align}
where the resulting random measure $G_0$ is a function of some covariates. 
Using $G_0$ as the base distribution of a DP for a document with a covariate, the random measure corresponding to document $m$ is constructed as follows:
\begin{align}
G_m &\sim \text{DP}(\beta, G_0(r_{m})),
\end{align}
where $G_0(r_{m})$ is the base distribution for the document with same covariate $r_{m}$ \citep{srebro2005time}. 

Similarly, the HDSP constructs a dependent random measure with covariates. However, unlike the DDP-DP approach, $G_0$ is no longer a function of covariates. The HDSP defines a single global random measure $G_0$ and then scales $G_0$ based on the covariates with the scaling function. The advantage of the HDSP is that it only requires a proper, but relatively simple, scaling function that reflects the correlation between covariates and topics, whereas the DDP requires a complex dependent process for different types of covariates \citep{RePEc:bes:jnlasa:v:101:y:2006:p:179-194}.

\section{\label{sec:vi}Variational Inference for HDSP}
The posterior inference for Bayesian nonparametric models is important because it is intractable to compute the posterior over an infinite dimensional space. Approximation algorithms, such as marginalized MCMC \citep{escobar1995bayesian, Teh:2006p3792} and variational inference \citep{blei2006variational, TehKurWel2008}, have been developed for the Bayesian nonparametric mixture models. We develop a mean field variational inference \citep{jordan1998introduction, wainwright2008graphical} algorithm for approximate posterior inference of the HDSP topic model. The objective of variational inference is to minimize the KL divergence between a distribution over the hidden variables and the true posterior, which is equivalent to maximizing the lower bound of the marginal log likelihood of observed data. 

In this section, we first derive the inference algorithm for the first scaling function with a fully factorized variational family. Variational inference algorithms can be easily modularized with the fully factorized variational family, and the variation in a model only affects the update rules for the modified parts of the model. Therefore, for the second scaling function, we only need to update the part of the inference algorithm related to the new scaling function.


\subsection{Variational inference for the first scaling function}
For the first scaling function, we use a fully factorized variational distribution and perform a mean-field variational inference. There are five latent variables of interest: the corpus level stick proportion $V_k$, the document level stick proportion $\pi_{mk}$, the scaling parameter between topic and label $w_{kj}$, the topic assignment for each word $z_{mn}$, and the word topic distribution $\phi_k$. Thus the variational distribution $q(z, \pi, V, w, \phi)$ can be factorized into
\begin{align}
&q(z, \pi, V, w, \phi) = \notag\\
&\prod_{k=1}^{T}\prod_{m=1}^{M}\prod_{j=1}^{J}\prod_{n=1}^{N_m} q(z_{mn}) q(\pi_{mk}) q(V_{k}) q(\phi_k) q(w_{kj}),
\end{align}
where the variational distributions are 
\begin{align}
&q(z_{mn}) = \text{Multinomial}(z_{mn} | \gamma_{mn}) &\notag\\
&q(\pi_{mk}) = \text{Gamma}(\pi_{mk} | a^\pi_{mk}, b^\pi_{mk}) &\notag\\
&q(V_k) = \delta_{V_k} &\notag\\
&q(\phi_k) = \text{Dirichlet}(\phi_k | \eta_{k}) &\notag\\
&q(w_{kj}) = \text{InvGamma}(w_{kj} | a^w_{kj}, b^w_{kj}). &\notag
\end{align}
For the corpus level stick proportion $V_k$, we use the delta function as a variational distribution for simplicity and tractability in inference steps as demonstrated in \citep{liang2007infinite}. Infinite dimensions over the posterior is a key problem in Bayesian nonparametric models and requires an approximation method. In variational treatment, we truncate the unbounded dimensionality to $T$ by letting $V_T = 1$. Thus the model still keeps the infinite dimensionality while allowing approximation to be carried out under the bounded variational distributions.

Using standard variational theory, we derive the evidence lower bound (ELBO) of the marginal log likelihood of the observed data $\mathcal{D} = (\mathbf{x}_m, \mathbf{r}_m)_{m=1}^M$,
\begin{align}
\label{eqn:elbo_hdsp}
&\log p(\mathcal{D}| \alpha, \beta, a^w, b^w, \eta) \notag\\
& \quad \geq \mathbb{E}_q[\log p(\mathcal{D}, z, \pi, V, w, \phi)] + H(q) = \mathcal{L}(q),
\end{align}
where $H(q)$ is the entropy for the variational distribution. By taking the derivative of this lower bound, we derive the following coordinate ascent algorithm.

\textbf{Document-level Updates:} At the document level, we update the variational distribution for the topic assignment $z_{mn}$ and the document level stick proportion $\pi_{mk}$.
The update for $q(z_{mn}|\gamma_{mn})$ is
\begin{align}
\gamma_{mnk} \propto \exp \left( \mathbb{E}_q[\ln \eta_{k,x_{mn}}] + \mathbb{E}_q[\ln \pi_{mk}] \right).
\end{align}
Updating $q(\pi_{mk}|a^\pi_{mk}, b^\pi_{mk})$ requires computing the expectation term $\mathbb{E}[\ln \sum_{k=1}^T \pi_{mk}]$. Following \cite{blei2007correlated}, we approximate the lower bound of the expectation by using the first-order Taylor expansion,
\begin{align}
-\mathbb{E}_q[\ln \sum_{k=1}^T \pi_{mk}] \geq - \ln \xi_m - \frac{\sum_{k=1}^{T}\mathbb{E}_q[\pi_{mk}] - \xi_m}{\xi_m},
\end{align}
where the update for $\xi_m = \sum_{k=1}^{K} \mathbb{E}_q[\pi_{mk}]$. Then, the update for $\pi_{mk}$ is
\begin{align}
a^\pi_{mk} &= \beta p_k + \sum_{n=1}^{N_m} \gamma_{mnk} \notag\\
b^\pi_{mk} &= \prod_j \mathbb{E}_q[w_{kj}^{-r_{mj}}] + \frac{N_m}{\xi_m}.
\end{align}
Note again $r_{mj}$ is equal to 1 when $j$th label is observed in $m$th document, otherwise 0.

\textbf{Corpus-level Updates:} At the corpus level, we update the variational distribution for the scaling parameter $w_{kj}$, corpus level stick length $V_k$ and word topic distribution $\eta_{ki}$.

The optimal form of a variational distribution can be obtained by exponentiating the variational lower bound with all expectations except the parameter of interest \citep{bishop2006pattern}. For $w_{kj}$, we can derive the optimal form of variational distribution as follows
\begin{align}
&q(w_{kj}) \sim \text{InvGamma}(a', b') \\
a' &= \mathbb{E}_{q}[ \beta p_k ] \sum_m r_{mj} + a^w \notag\\
b' &=  \sum_{m'}\prod_{j'/j}\mathbb{E}_q[w_{j'k}^{-1}] \mathbb{E}_q[ \pi_{m'k}] + b^w ,\notag
\end{align}
where $m' = \{m:r_{mj}=1\}$ and $j'/j$ $=$ $\{j':r_{mj'} = 1, j' \neq j\}$. See Appendix A for the complete derivation.
There is no closed form update for $V_k$, instead we use the steepest ascent algorithm to jointly optimize $V_k$. The gradient of $V_k$ is
\begin{align}
&\frac{\partial \mathcal L}{\partial V_k} = - \frac{\alpha-1}{1-V_k} &\\
&- \frac{\beta p_k}{V_k}\{\sum_{m,j} r_{mj} \mathbb{E}_q[\ln \pi_{mk}] - \mathbb{E}_q[\pi_{mk}] + \psi(\beta p_k) \} & \notag\\
&+ \sum_{k'>k}\frac{\beta p_{k'}}{1-V_k}\{\sum_{m,j} r_{mj} \mathbb{E}_q[\ln \pi_{mk'}] - \mathbb{E}_q[\pi_{mk'}] + \psi(\beta p_{k'})\},  & \notag
\end{align}
where $\psi(\cdot)$ is a digamma function. Finally, the update for the word topic distribution $q(\phi_k|\eta_k)$ is 
\begin{align}
\eta_{ki} = \eta + \sum_{m,n} \gamma_{mnk} \mathbf{1}(\mathrm{x}_{mn} = i),
\end{align}
where $i$ is a word index, and $\mathbf{1}$ is an indicator function \citep{Blei:2003p4796}.

The expectations of latent variables under the variational distribution $q$ are
\begin{align}
&\mathbb{E}_q[\pi_{mk}] = a^\pi_{mk}/b^\pi_{mk} &\notag\\
&\mathbb{E}_q[\ln \pi_{mk}] = \psi(a^\pi_{mk}) - \ln b^\pi_{mk} &\notag\\
&\mathbb{E}_q[w_{kj}] = b^w_{kj}/(a^w_{kj}-1)&\notag\\
&\mathbb{E}_q[w_{kj}^{-1}] = a^w_{kj}/b^w_{kj}&\notag\\
&\mathbb{E}_q[\ln w_{kj}] = \ln b^w_{kj} - \psi(a^w_{kj})&\notag\\
&\mathbb{E}_q[\ln \phi_{ki}] = \psi(\eta_{ki}) - \psi(\sum_i \eta_{ki}).&\notag
\end{align}
\subsection{Variational inference for the second scaling function}
Introducing a new scaling function requires a new approximation method. We first choose the part of ELBO which requires new treatment as the scaling function changes. From Equation \ref{eqn:elbo_hdsp}, we take the terms that are related to the scaling function $s$:
\begin{align}
\mathcal L_s & =\mathbb{E}_q[\sum_{m=1}^{M} \sum_{k=1}^{\infty} \ln p(\pi_{mk}|V_k, s, \mathbf{r}_m)] + \mathbb{E}_q[ \ln p(s)] - \mathbb{E}_q[\ln q(s)] \\
& = \sum_m \left[ \beta p_k  \mathbb{E}_q[ \ln s(r_m) ] + (\beta p_k -1) \mathbb{E}_q[\ln \pi_{mk}] - \mathbb{E}_q[s(r_m)] \mathbb{E}_q[\pi_{mk}] - \ln\Gamma(\beta p_k)  \right] \notag\\
& \quad + \mathbb{E}_q[p(s)] - \mathbb{E}_q[q(s)]. \notag
\end{align}
To update the scaling parameters, we need a proper prior and variational distribution. For the second scaling function, the normal distribution with zero mean and variance $\sigma$ is used as a prior of $w_{kj}$, and the delta function is used as the variational distribution of $w_{kj}$. Newton-Raphson optimization method are used to update the weight parameters. The Newton-Raphson optimization finds a stationary point of a function by iterating:
\begin{align}
w_{k}^{\text{new}} \leftarrow w_{k}^{\text{old}} - H({w_k})^{-1} \frac{\partial{\mathcal L}}{\partial{w_{k}}},
\end{align}
where $H(w_k)$ and $\frac{\partial{\ell}}{\partial{w_{k}}}$ are the Hessian matrix and gradient at the point $w_{k}^{\text{old}}$. The lower bound with respect to the parameter $w_{kj}$ is,
\begin{align}
\mathcal L_{w_{kj}} = \sum_m \left[ \beta p_k \sum_j w_{kj}r_{mj} + (\beta p_k -1) \mathbb{E}_q[\ln \pi_{mk}] - \exp(w^{\top}_k r_m) \mathbb{E}_q[\pi_{mk}] - \ln\Gamma(\beta p_k)  \right].
\end{align}
Then, the gradient and Hessian matrix of $w_{kj}$ are 
\begin{align}
\frac{\partial{\mathcal L}}{\partial{w_{kj}}} &= \sum_m \left[\beta p_k r_{mj} - r_{mj} \exp(w^{\top}_k r_{m}) \mathbb{E}_q[\pi_{mk}] \right] - \frac{w_{jk}}{\sigma}	\\
\frac{\partial^2{\mathcal L}}{\partial{w_{kj'}}\partial{w_{kj}}} &= \sum_m \left[ - r_{mj'} r_{mj} \exp(w^{\top}_k r_{m}) \mathbb{E}_q[\pi_{mk}]\right]  - \mathbf{1}(j=j')\sigma^{-1}.
\end{align}
Because the $w_{kj}$ is depends on both label $j$ and topic $k$, we iteratively update $w_{kj}$ until converged.


The update rules for the variational parameter of $\pi_{mk}$ and $V_k$ need to be changed to accommodate the change of scaling function. The variational parameters of $\pi_{mk}$ are approximated by using the first-order Taylor expansion,
\begin{align}
\label{pi_update}
a^{\pi}_{mk} = \beta p_k + \sum_{n=1}^{N_m} \gamma_{mnk} \\ 
b^{\pi}_{mk} = \mathbb{E}_q[\ln s(r_{m})] + \frac{N_m}{\xi_m},	\notag
\end{align}
where $\xi_m$ is $\sum_{k=1}^{K} \mathbb{E}_q[\pi_{mk}]$. To update $V_k$, we take the same approach in which the variational distribution is a delta function of current $V_k$. Again, we use the steepest ascent algorithm to jointly optimize $V_k$, and the gradient of $V_k$ is
\begin{align}
\label{v_update}
&\frac{\partial \mathcal L}{\partial V_k} = - \frac{\alpha-1}{1-V_k} - \frac{\beta p_k}{V_k}\left\{\sum_{m}  \mathbb{E}_q[\ln s(r_m)] \mathbb{E}_q[\ln \pi_{mk}] - \mathbb{E}_q[\pi_{mk}] + \psi(\beta p_k) \right\} & \\
&+ \sum_{k'>k}\frac{\beta p_{k'}}{1-V_k}\left\{\sum_{m}  \mathbb{E}_q[\ln s(r_m)] \mathbb{E}_q[\ln \pi_{mk'}] - \mathbb{E}_q[\pi_{mk'}] + \psi(\beta p_{k'})\right\}.  & \notag
\end{align}
The update rules for $\pi_{mk}$ and $V_k$ only requires the expectation of the log scaling function. The update rules for the other parameters remain the same as the previous section.

Introducing a new scaling function requires a new inference algorithm, and this can be cumbersome. Once one defines a tractable expectation of the log of a scaling function, a recently suggested Black-box method \citep{ranganath2014black} can be an alternative to update the function-specific parameters instead of deriving function-specific approximation algorithms.

\section{\label{sec:exp}Experiments}


In this section, we describe how the HDSP performs with real and synthetic data. We fit the HDSP topic model with three different types of data and compare the results with several comparison models. First, we test the model with synthetic data to verify the approximate inference. Second, we train the model with categorical data whose label information is represented by binary values. Third, we train the model with mixed-type of data whose label information has both numerical and categorical values.


\subsection{Synthetic data}
There is no naturally-occurring dataset with the observable weights between topics and labels, so we synthesize data based on the model assumptions to verify our model and the approximate inference. First, we check the difference between the original topics and the inferred topics via simple visualization. Then, we focus on the differences between the inferred and synthetic weights. For all experiments with synthetic data, the datasets are generated by following the model assumptions with the first scaling function, and the posterior inferences are done with the first scaling function. We set the truncation level $T$ at twice the number of topics. We terminate the variational inference when the fractional change of the lower bound falls below $10^{-3}$, and we average all results over 10 individual runs with different initializations.

\begin{figure}[t!]
	\centering

	\subfigure[Synthetic Topics\label{fig:syn_topic}]{
	\includegraphics[width=0.3\linewidth, trim=15 30 50 40, clip=true]{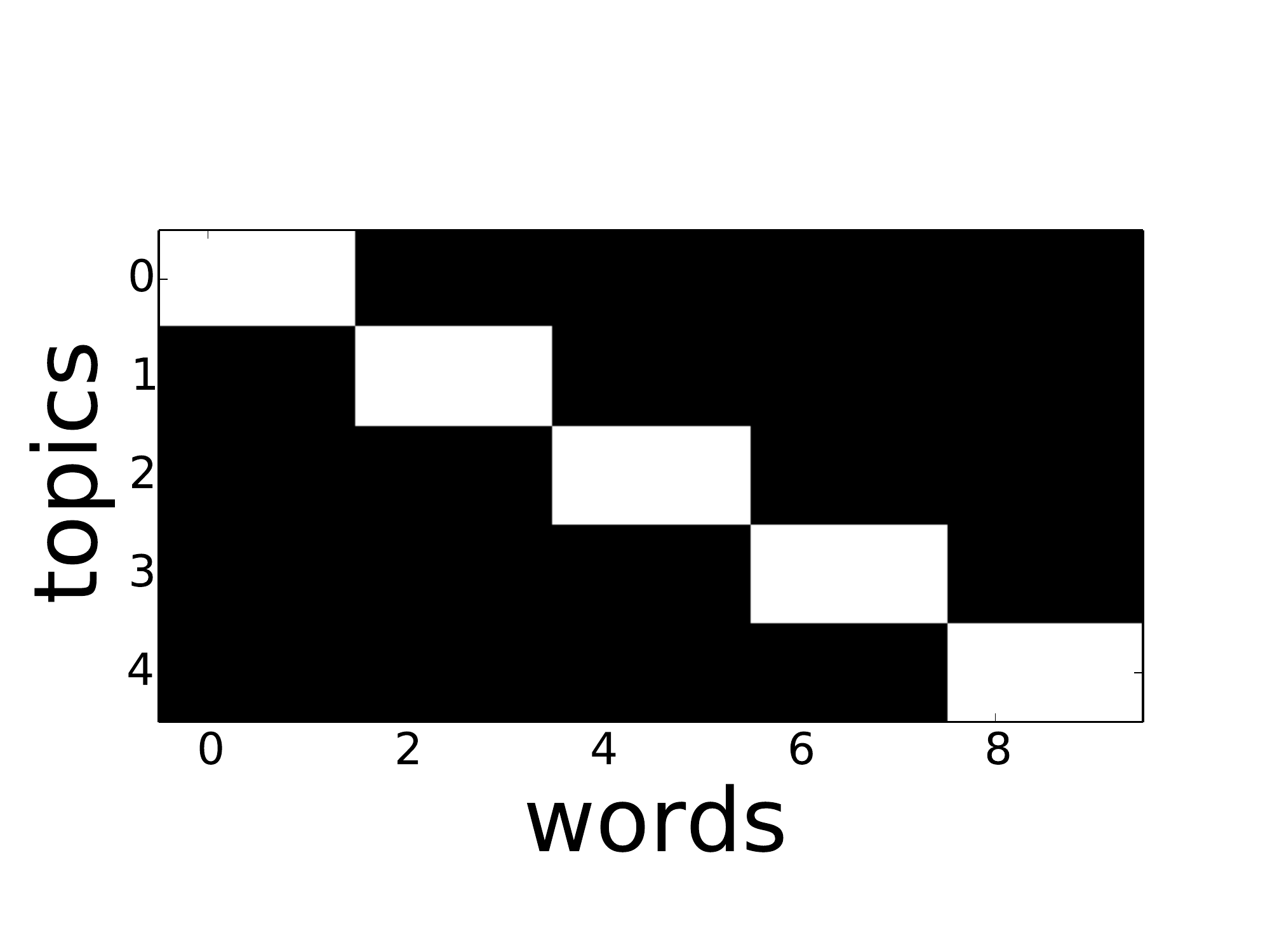}			
	}
	\subfigure[HDSP Topics \label{fig:hdsp_topic}]{
	\includegraphics[width=0.3\linewidth, trim=15 30 50 40, clip=true]{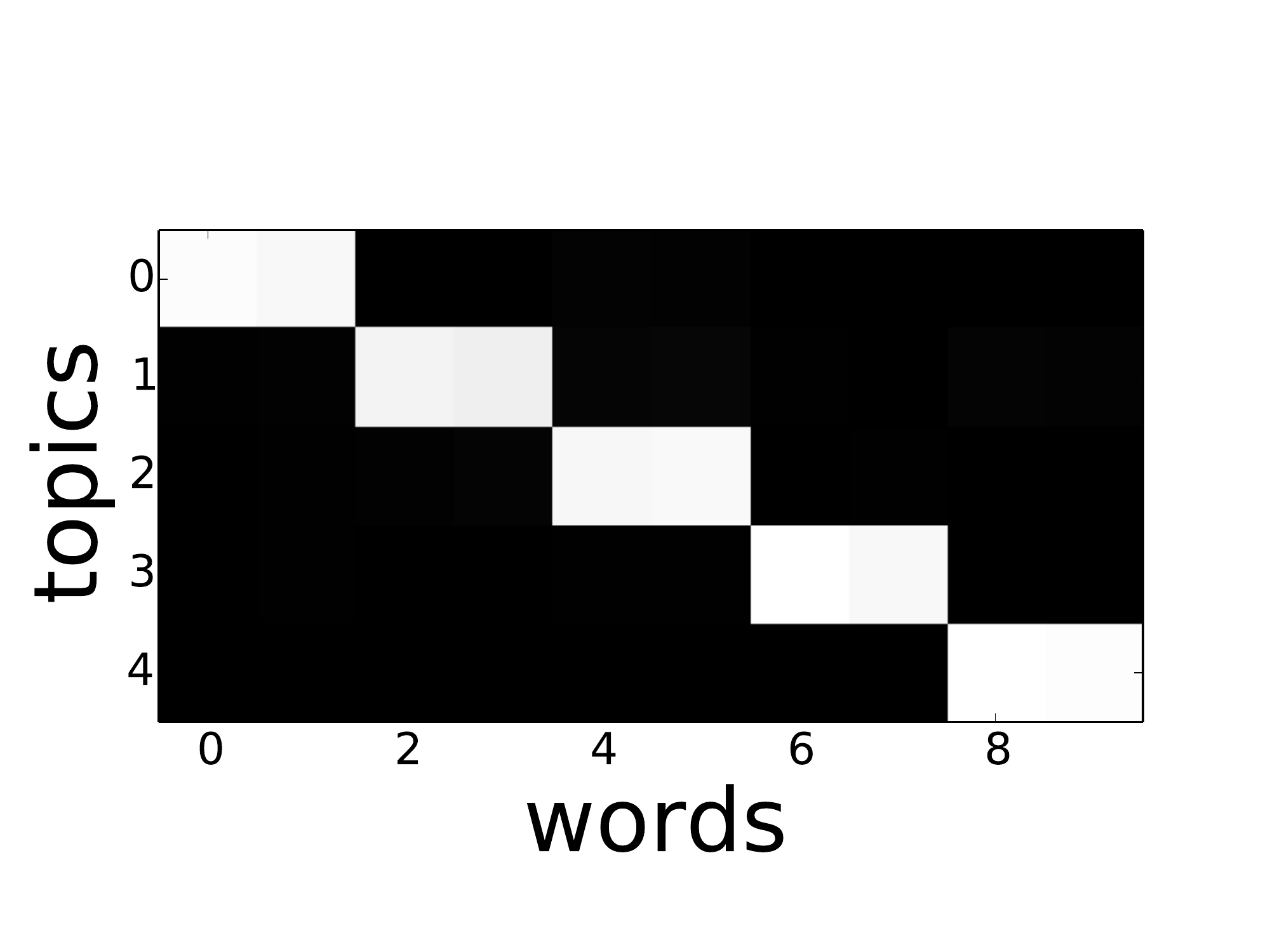}			
	}
	\subfigure[HDP Topics \label{fig:hdp_topic}]{
	\includegraphics[width=0.3\linewidth, trim=15 30 50 40, clip=true]{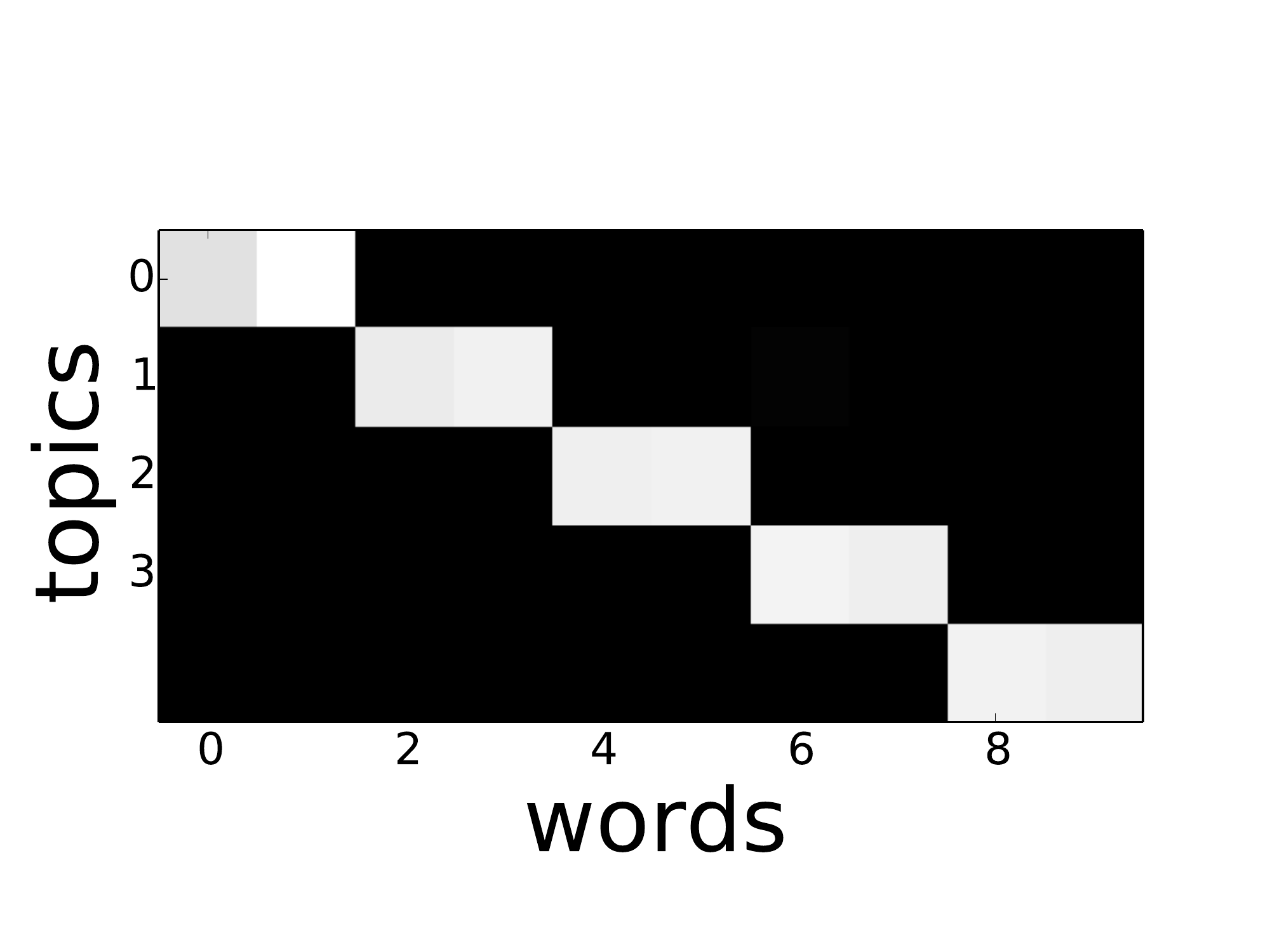}			
	}
	\\
	\subfigure[Syn scaling parameter\label{fig:syn_rloc}]{	
	\includegraphics[width=0.3\linewidth, clip=true]{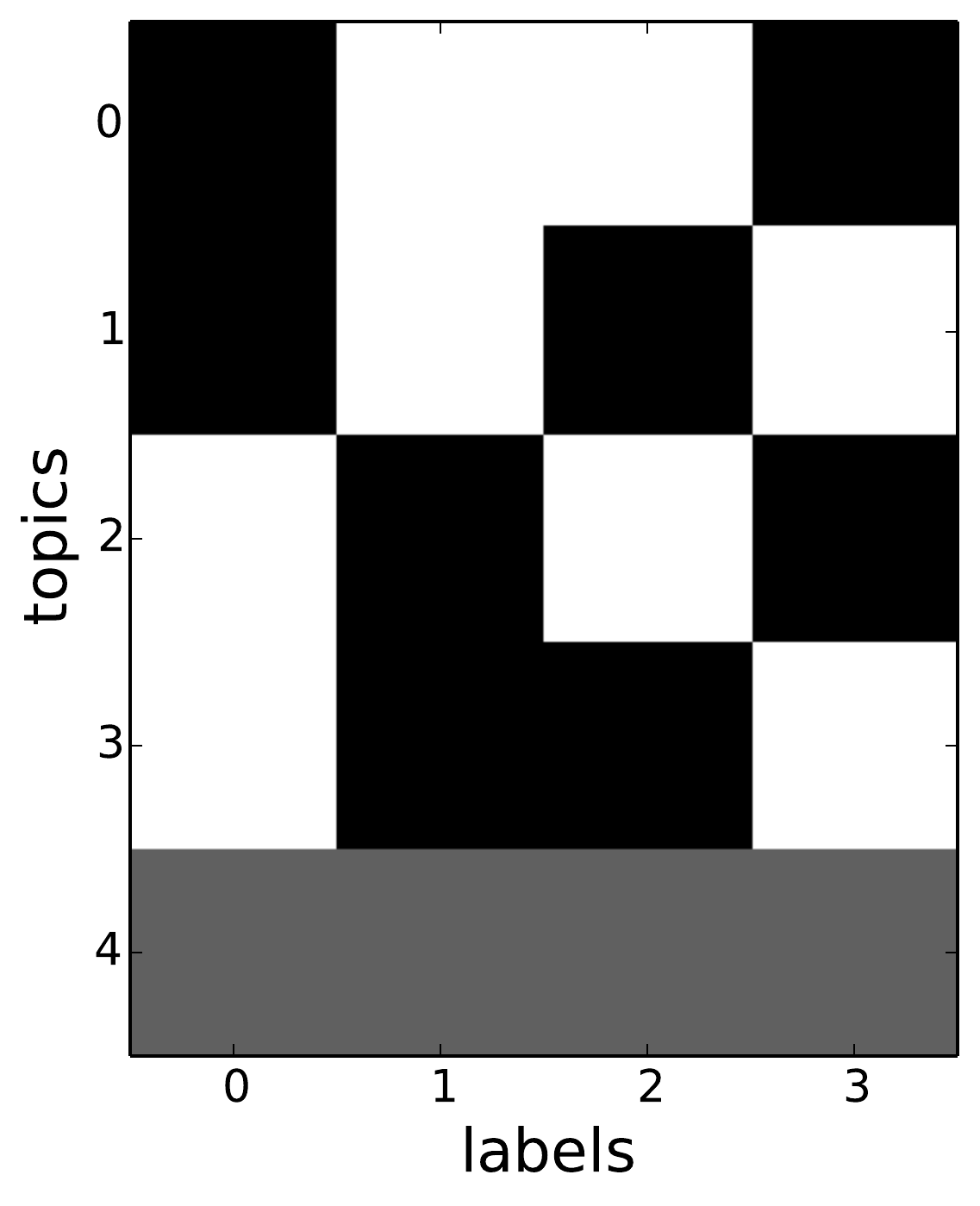}		
	}
	\subfigure[HDSP scaling parameter\label{fig:inf_rloc}]{
	\includegraphics[width=0.3\linewidth, clip=true]{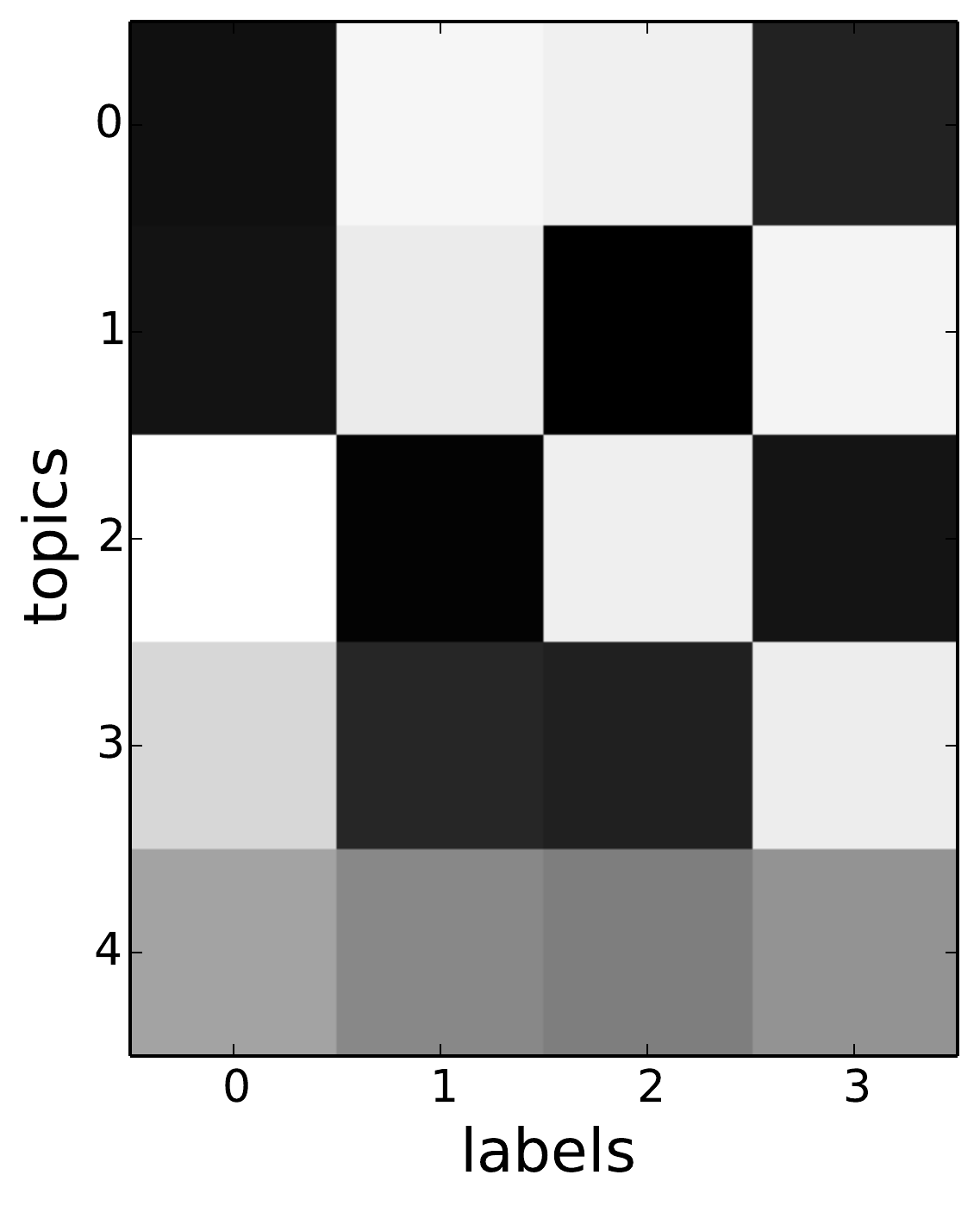}		
	}
		
	\caption{\label{fig:synhdsp}Experiments with synthetic data. (a) is the synthetic topic distribution of 5 topics over 10 terms. (b) and (c) are topic distributions inferred by the HDSP and the HDP. Both models recover the original topics. 
	(d) shows the heat map of original scaling parameters between the topics and labels. (e) shows the heat map of the recovered parameters by HDSP.}
\end{figure}

\begin{figure}[ht!]
	\centering
	\includegraphics[width=0.48\linewidth, trim=10 10 10 10, clip=true]{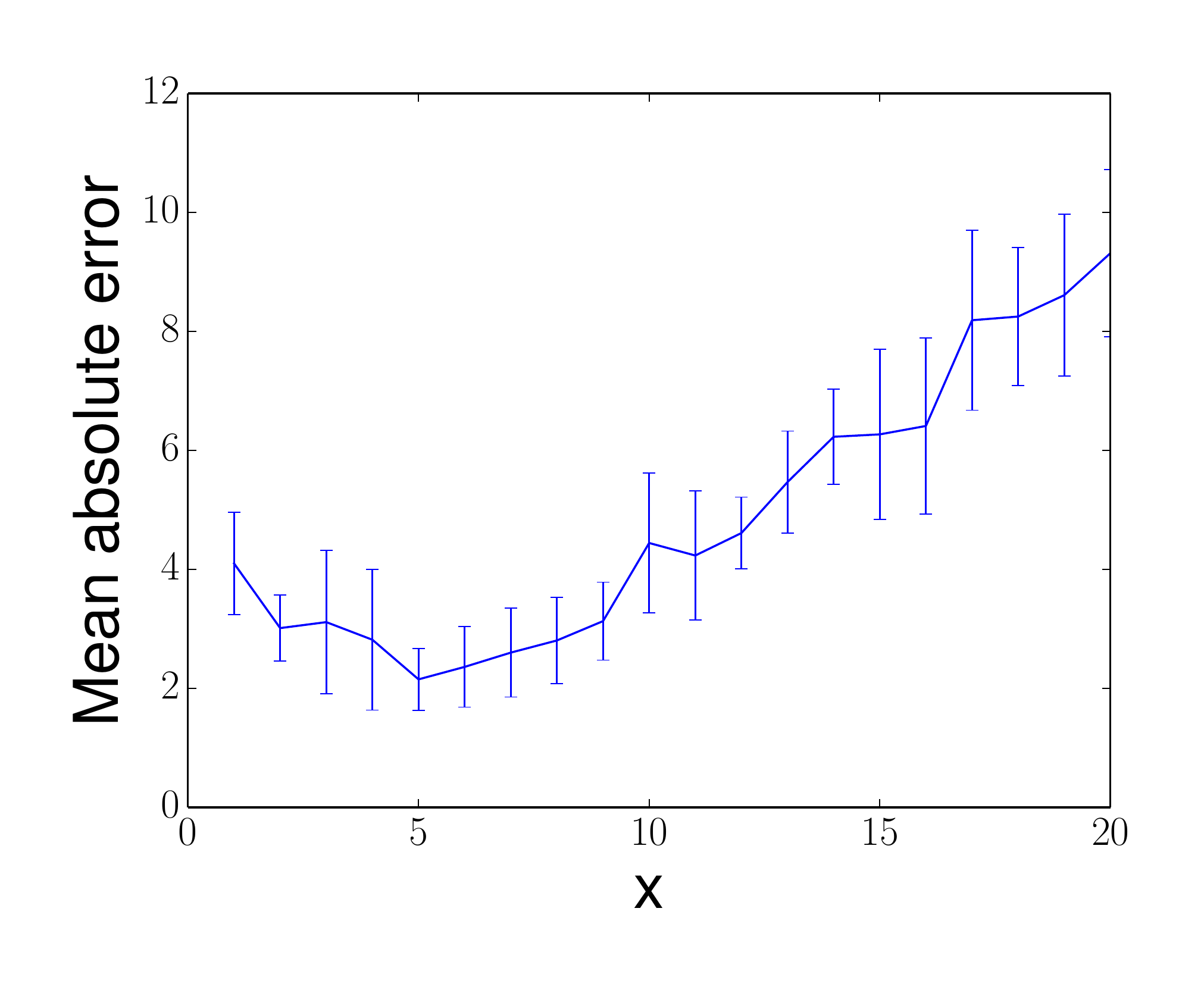}
	\includegraphics[width=0.48\linewidth, trim=10 10 10 10, clip=true]{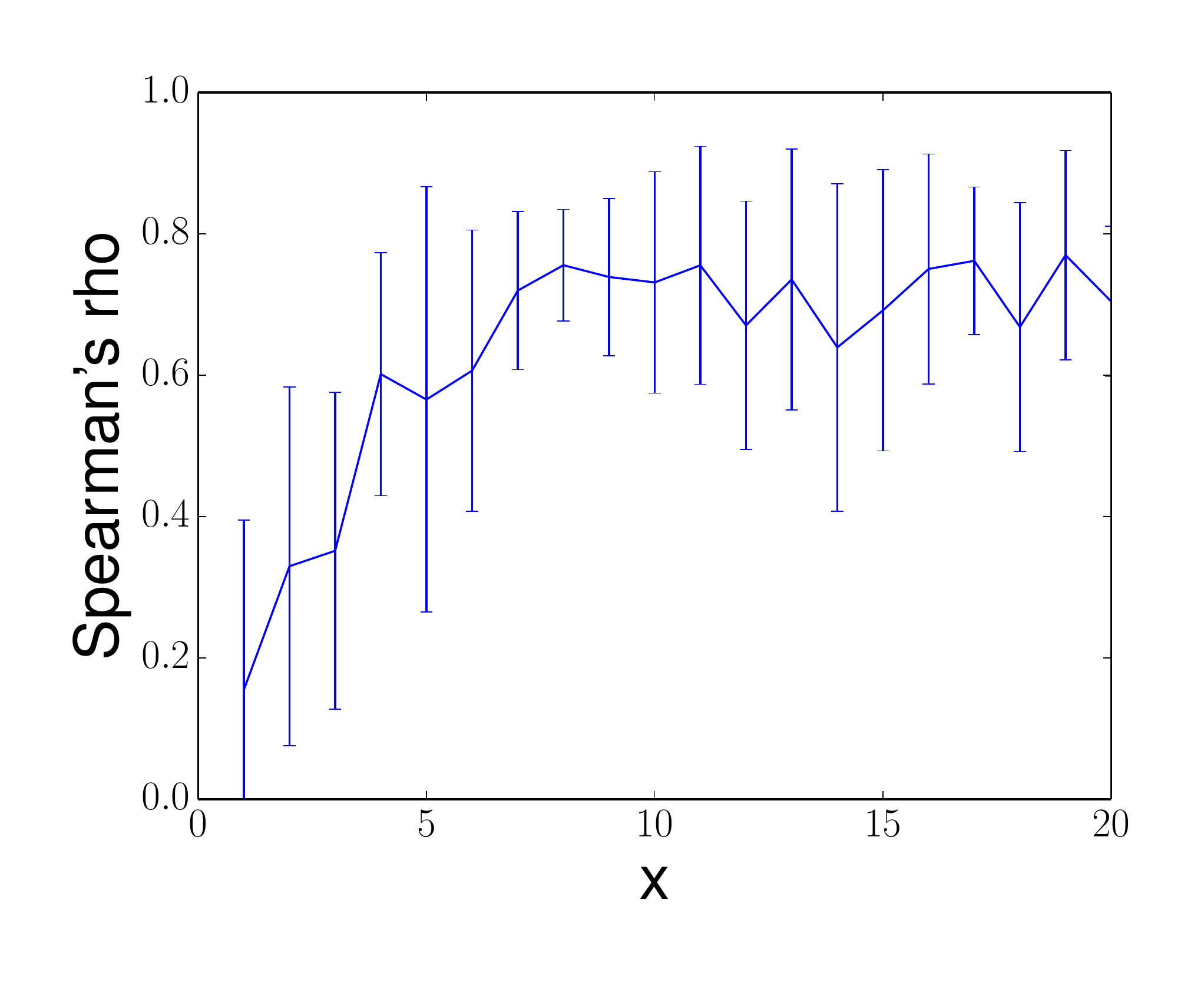}			
	\caption{\label{fig:syn_spear} Spearman's correlation coefficient and mean absolute error of the synthetic data with various volume of space ($x^3$). As the volume of space for locations increases, the mean absolute error also increases (left). However, the model preserves the relative weights between topics and labels, shown by the high and stabilized correlation between the original ordering and the recovered ordering of label-topic pairs in terms of the weights between the two (right). This is a key characteristic of the HDSP model which scales the mixture components according to the inverse of the weights.}
\end{figure}

With the first experiment, we show that HDSP correctly recovers the underlying topics and scaling parameter between topics and labels. For the dataset, we generate 500 documents using the following steps. We define five topics over ten terms  shown in Figure \ref{fig:syn_topic} and the scaling parameter of five topics and four labels shown in Figure \ref{fig:syn_rloc}.  For each document, we randomly draw $N_m$ from the Poisson distribution and $r_{mj}$ from the Bernoulli distribution. The average length of a document is 20, and the average number of labels per document is 2.  We generate topic proportions of corpus and documents by using Equations \ref{eqn:st1} and \ref{eqn:st4}. For each word in a document, we draw the topic and the word by using Equation \ref{eqn:mmt}. We set both $\alpha$ and $\beta$ to 1.

Figure \ref{fig:synhdsp} shows the results of the HDP and the HDSP on the synthetic dataset. Figure \ref{fig:hdsp_topic} and Figure \ref{fig:hdp_topic} are the heat maps of topics inferred from each model. We match the inferred topics to the original topics using KL divergence between the two sets of topic distributions. There are no significant differences between the inferred topics of HDSP and HDP. In addition to the topics, HDSP infers the scaling parameters between topics and labels, which are shown in Figure \ref{fig:inf_rloc}. The results show that the relative differences between original scaling parameters are preserved in the inferred parameters through the variational inference.

With the second experiment, we show that the inferred parameters preserve the relative differences between labels and topics in the dataset.  For this experiment, we generate 1,000 documents with ten randomly drawn topics from Dirichlet(0.1) with the vocabulary size of 20. To generate the weights between topics and labels, we randomly place the topics and labels into three dimensional euclidean space, and use the distance between a topic and label as a scaling parameter. The location of topics and labels are uniformly drawn from three dimensional euclidean space, so the total volume is $x^3$, then we vary the $x$ value from 1 to 20 for each experiment. As the volume of space increases, the potential scaling parameter between a topic and label increases, and the scaling effect on a topic proportion of document also increases.

We compute the mean absolute error (MAE) and the spearman's rank correlation coefficient $\rho$ between the original parameters and the inferred parameters. The spearman's $\rho$ is designed to measure the ranking correlation of two lists. Figure \ref{fig:syn_spear} shows the results. The MAE increases as the volume of the space increases. However, spearman's $\rho$ stabilizes, indicating that the relative differences are preserved even when the MAE increases. Since there are an infinite number of configurations of scaling parameters that generate the same expectation $\mathbb{E}[p(\pi_{m}|\beta p, w_j)]$ given $\pi_{m}$ and $\beta p$, preserving the relative differences verifies our model's capability of capturing the underlying structure of topics and labels.

\subsection{\label{real_result}Categorical data}
We evaluate the performance of HDSP and compare it with the HDP, labeled LDA (L-LDA), partially labeled LDA (PLDA), and author-topic model (ATM). For the HDSP, we use both scaling functions and denote the model with the second scaling function as wHDSP.
We use three multi-labeled corpora: RCV\footnote{http://trec.nist.gov/data/reuters/reuters.html} (newswire from Reuter's), OHSUMED\footnote{http://ir.ohsu.edu/ohsumed/ohsumed.html} (a subset of the Medline journal articles), and NIPS (proceedings of NIPS conference).
For RCV and OHSUMED, we use multi-category information of documents as labels, and for NIPS, we use authors of papers as labels. The average number of labels per article is 3.2 for RCV, 5.2 for OHSUMED, and 2.4 for NIPS. Table \ref{tbl:dataset3} contains the details of the datasets.

\begin{table}[t!]
\centering
\caption{\label{tbl:dataset3}Datasets used for the experiments in \ref{real_result}. As the last two columns show, we experiment on datasets with a varied number of unique labels, as well as the average number of labels per document. 
}
\begin{tabular}{l|rrrrr}
      & docs & vocab & labels & labels/doc & doc/labels\\\hline
RCV& 23,149 & 9,911 & 117   & 3.2 & 729.7\\
OHSUMED & 7,505 & 7,056 & 52 & 5.2 & 722.0\\
NIPS & 2,484 & 14,036 & 2,865 & 2.4 & 1.6
\end{tabular}
\end{table}

\subsubsection{Experimental settings}
For the HDP and HDSP, we initialize the word-topic distribution with three iterations of LDA for fast convergence to the posterior while preventing the posterior from falling into a local mode of LDA and then reorder these topics by the size of the posterior word count. For all experiments, we set the truncation level $T$ to 200. We terminate variational inference when the fractional change of the lower bound falls below $10^{-3}$, and we optimize all hyperparameters during inference except $\eta$. For the L-LDA and PLDA, we implement the collapsed Gibbs sampling algorithm. For each model, we run 5,000 iterations, the first 3,000 as burn-in and then using the samples thereafter with gaps of 100 iterations. For PLDA, we set the number of topics for each label to two and five (PLDA-2, PLDA-5). For the ATM, we set the number of topics to 50, 100, and 150. We try five different values for the topic Dirichlet parameter $\eta$: $\eta = 0.1, 0.25, 0.5, 0.75, 1.0$. Finally all results are averaged over 20 runs with different random initialization. We do not report the standard errors because they are small enough to ignore. 

\begin{figure*}[t!]
	\centering
	\subfigure[OHSUMED\label{fig:ll_ohsumed}]{
	\includegraphics[width=0.45\linewidth, clip=true]{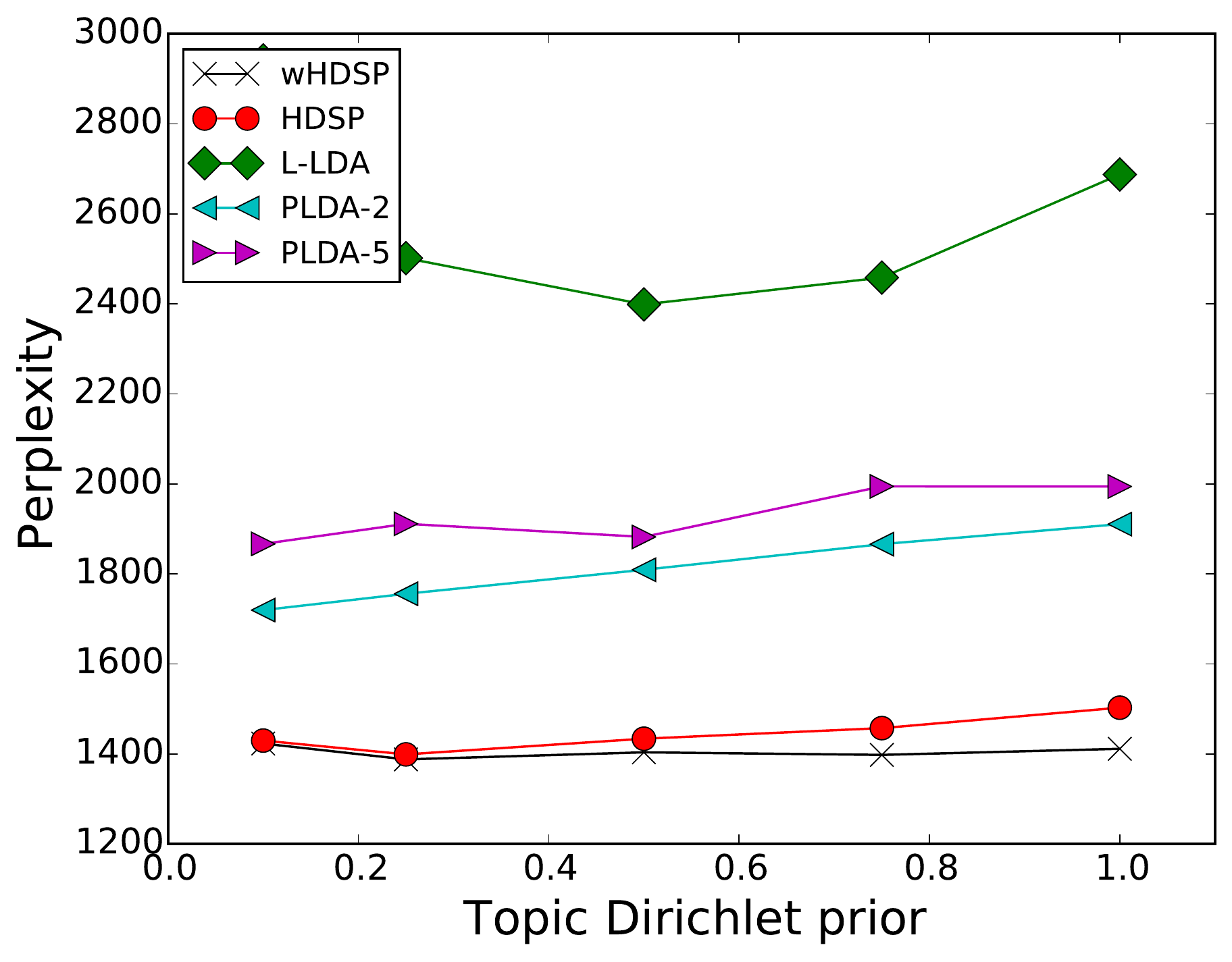}			
	}
	\subfigure[RCV\label{fig:ll_rcv}]{	
	\includegraphics[width=0.45\linewidth, clip=true]{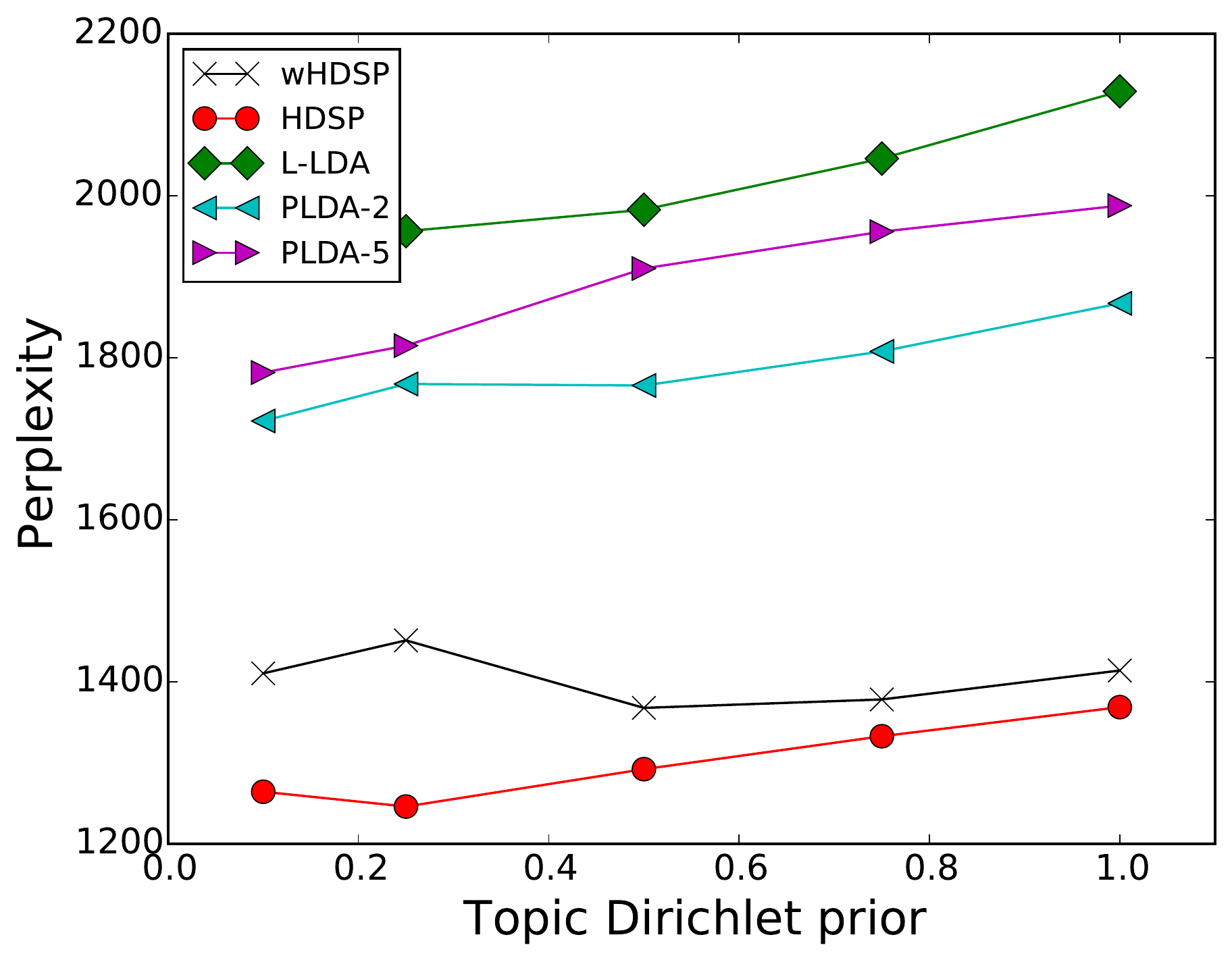}
	}
	\subfigure[NIPS\label{fig:ll_nips_hdsp}]{
	\includegraphics[width=0.45\linewidth, clip=true]{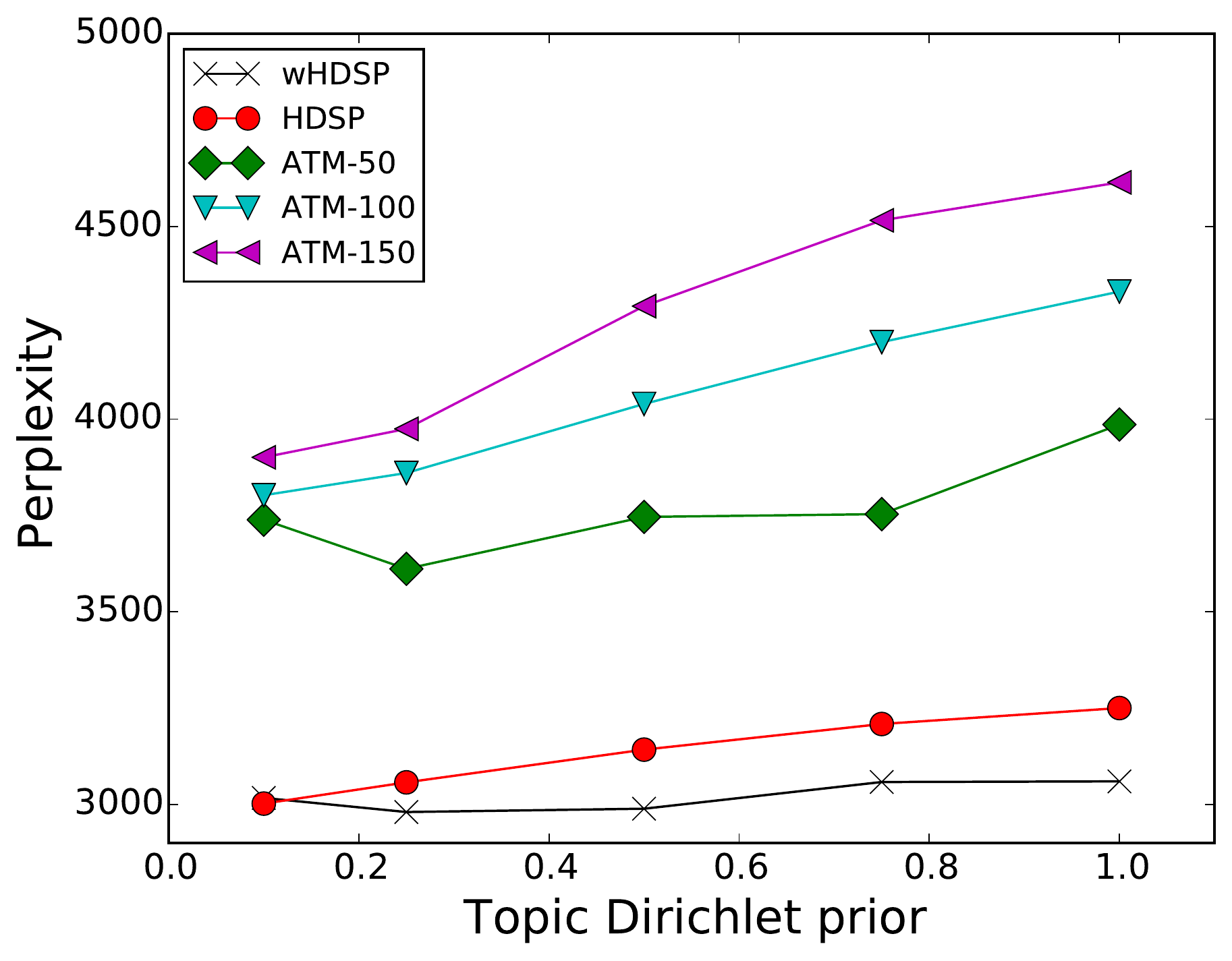}
	}
	\caption{\label{fig:heldout_ll} Perplexity of held-out documents. For HDSP (first scaling function), wHDSP (second scaling function), L-LDA, ATM, and PLDA, the perplexity is measured given documents and observed labels. For HDP, the model only uses the words of the documents. The HDSP which, instead of excluding topics from unobserved labels, scales all topics according to observed labels, shows the best heldout perplexity.}
\end{figure*}

\subsubsection{Evaluation metric}
The goal of our model is to construct the dependent random probability measure given multiple labels. Therefore, our interest is to see the increments of predictive performance when the label information is given.

The predictive probability given label information for held-out documents are approximated by the conditional marginal,
\begin{align}
&p(\mathbf{x}'|\mathbf{r}', \mathcal{D}_{\text{train}}) = &\\
&\int_q \prod_{n=1}^{N} \sum_{k=1}^T p(\mathrm{x}_n'|\phi_k) p(z_n'=k|\pi') p(\pi'|V, \mathbf{r}') d q(V, w, \phi),& \notag
\end{align}
where $\mathcal{D}_{\text{train}} = \{\mathbf{x}_{\text{train}}, \mathbf{r}_{\text{train}}\}$ is the training data, $\mathbf{x}'$ is the vector of $N$ words of a held-out document, $\mathbf{r}'$ are the labels of the held-out document, $z_n'$ is the latent topic of word $n$, and $\pi_k'$ is the $k$th topic proportion of the held-out document. Since the integral is intractable, we approximate the probability 
\begin{align}
p(\mathbf{x}'|\mathbf{r}', \mathcal{D}_{\text{train}}) \approx \prod_{n=1}^N\sum_{k=1}^T \tilde\pi_k \tilde\phi_{k, \mathrm{x}_n'},
\end{align}
where $\tilde\phi_{k}$ and $\tilde\pi_k$ are the variational expectations of $\phi_k$ and $\pi_k$ given label $\mathbf{r}'$.
This approximated likelihood is then used to compute the perplexity of the held-out document
\begin{align}
\label{eqn:perplexity_hdsp}
\text{perplexity} = \exp \left\{ \frac{-\ln p(\mathbf{x}' |\mathbf{r}', \mathcal{D}_{\text{train}})}{N} \right\}.
\end{align}
Lower perplexity indicates better performance. We also take the same approach to compute the perplexity for L-LDA, PLDA and HDP, but HDP does not use the labels of held-out documents. To measure the predictive performance, we leave 20\% of the documents for testing and use the remaining 80\% to train the models.


\subsubsection{Experimental results}
Figure \ref{fig:heldout_ll} shows the predictive performance of our model against the comparison models. For the OHSUMED and RCV corpora, both HDSP and wHDSP outperform all others. Among these models, L-LDA restricts the modeling flexibility the most; the PLDA relaxes that restriction by adding an additional latent label and allowing multiple topics per label. HDSP and wHDSP further increase the modeling flexibility by allowing all topics to be generated from each label. This is reflected in the results of predictive performance of the three models; L-LDA shows the worst performance, then PLDA, and HDSP and wHDSP show the lowest perplexity. 
For the NIPS data, we compare HDSP and wHDSP to ATM, and again, HDSP and wHDSP show the lowest perplexity. 

\begin{figure}[t!]
	\centering
	\includegraphics[width=0.8\linewidth]{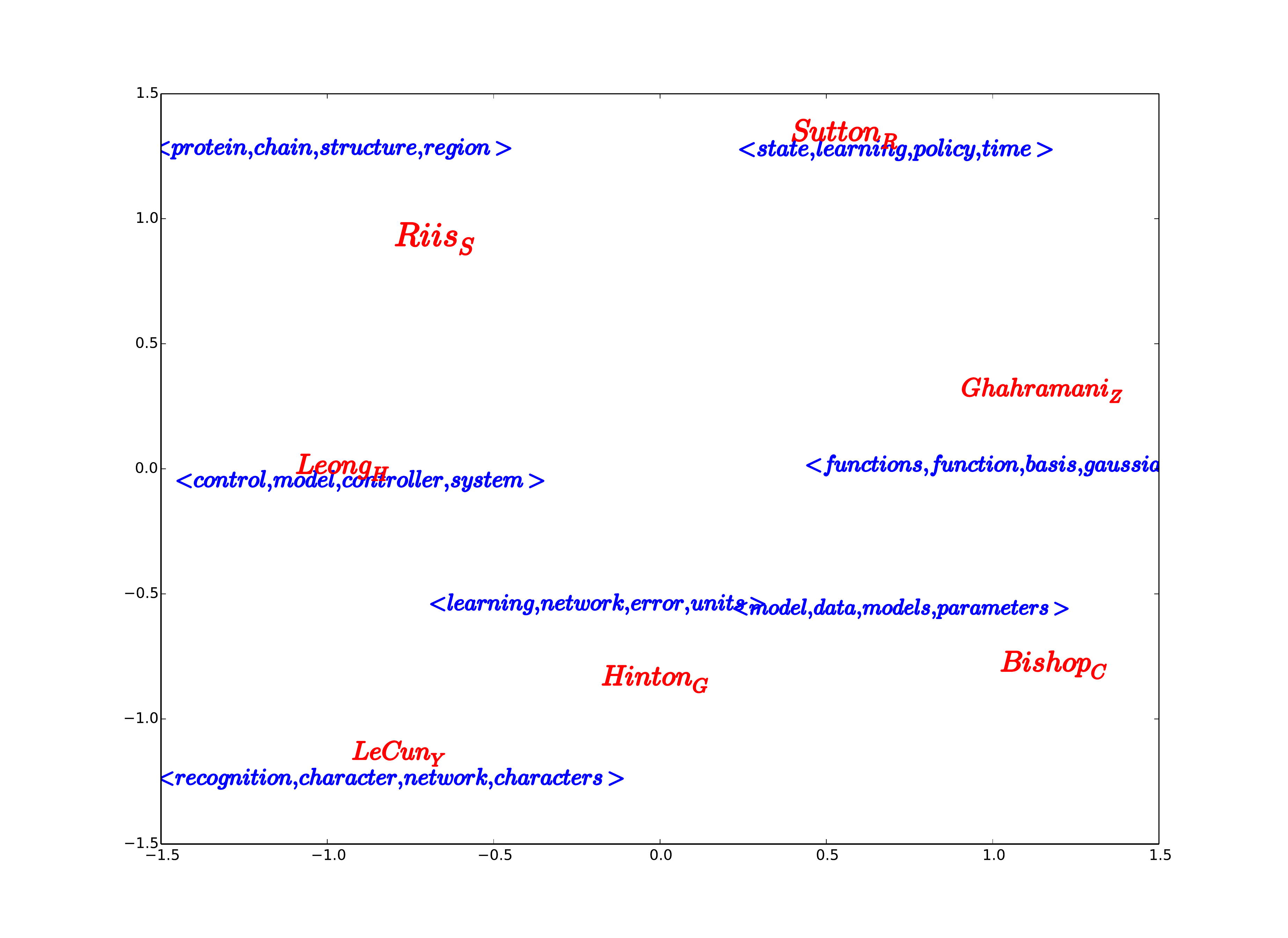}
	\caption{\label{fig:location_nips}Relative locations of observed labels (red) and latent topics (blue) inferred by HDSP from the NIPS corpus}	
\end{figure}	
\begin{figure}[t!]	
	\centering
	\includegraphics[width=0.8\linewidth]{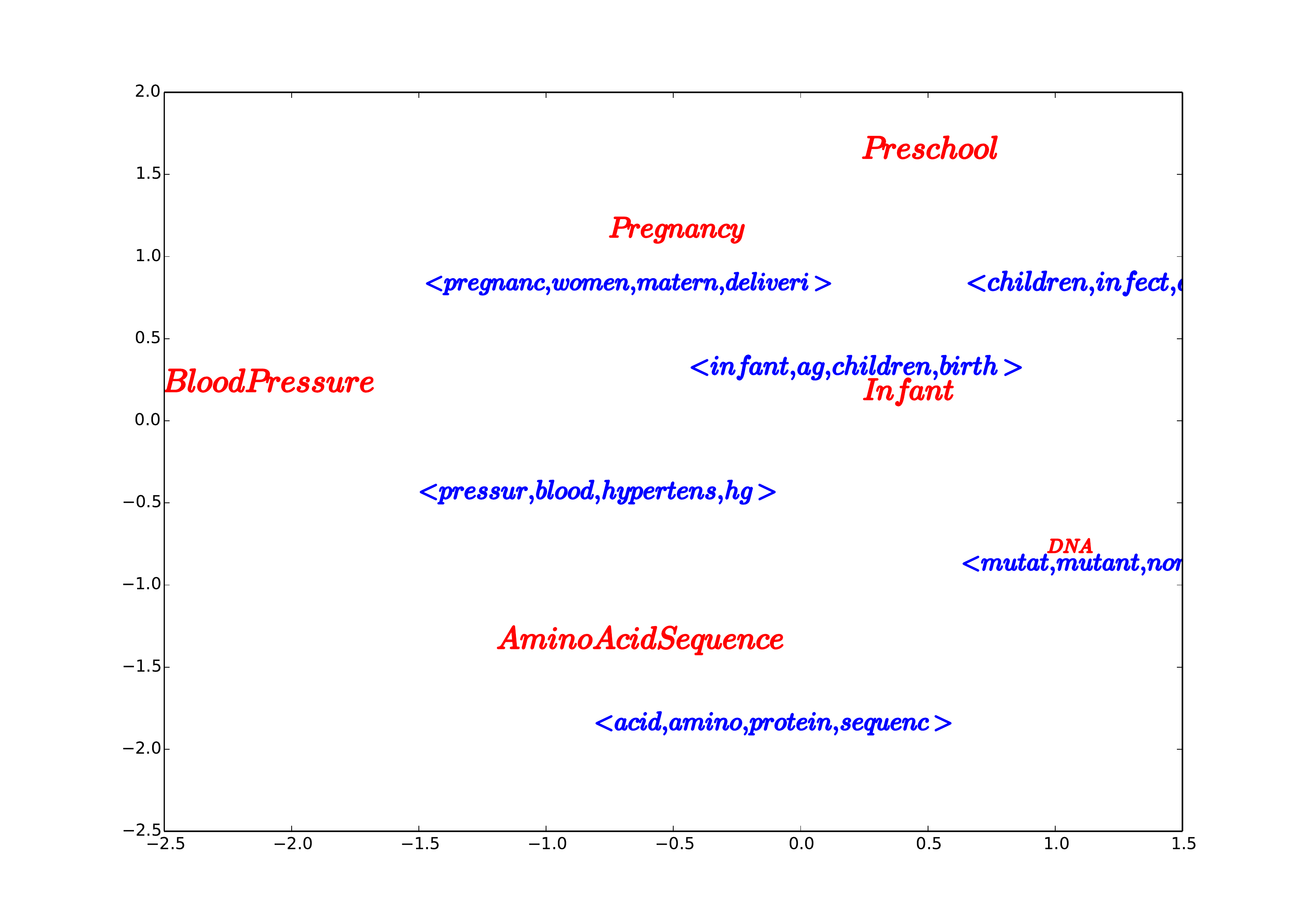}	
	\caption{\label{fig:location_ohsumed}Relative locations of observed labels (red) and latent topics (blue) inferred by HDSP from the OHSUMED corpus.}	
\end{figure}

To visualize the relationship between topics and labels, we embed the inferred topics and the labels into the two dimensional euclidean space by using multidimensional scaling  \citep{kruskal1964multidimensional} on the inferred parameters of HDSP. In Figure \ref{fig:location_nips}, we choose and display a few representative topics and authors from NIPS. For instance, Geoffrey Hinton and Yann LeCun are closely located to the neural network related topics such as `learning, network error' and `recognition character, network', and the reinforcement learning researcher Richard Sutton is closely located to the `state, learning policy' topic. Figure \ref{fig:location_ohsumed} shows the embedded labels and topics from OHSUMED. The labels `Preschool', `Pregnancy', and `Infant' are closely located to one another with similar topics. While the model explicitly models the correlation between topics and labels, embedding them together shows that the correlation among labels, as well as among topics, can also be inferred.

\begin{figure*}[h!]
	\centering
	\subfigure[RCV]{
	\includegraphics[width=0.96\linewidth]{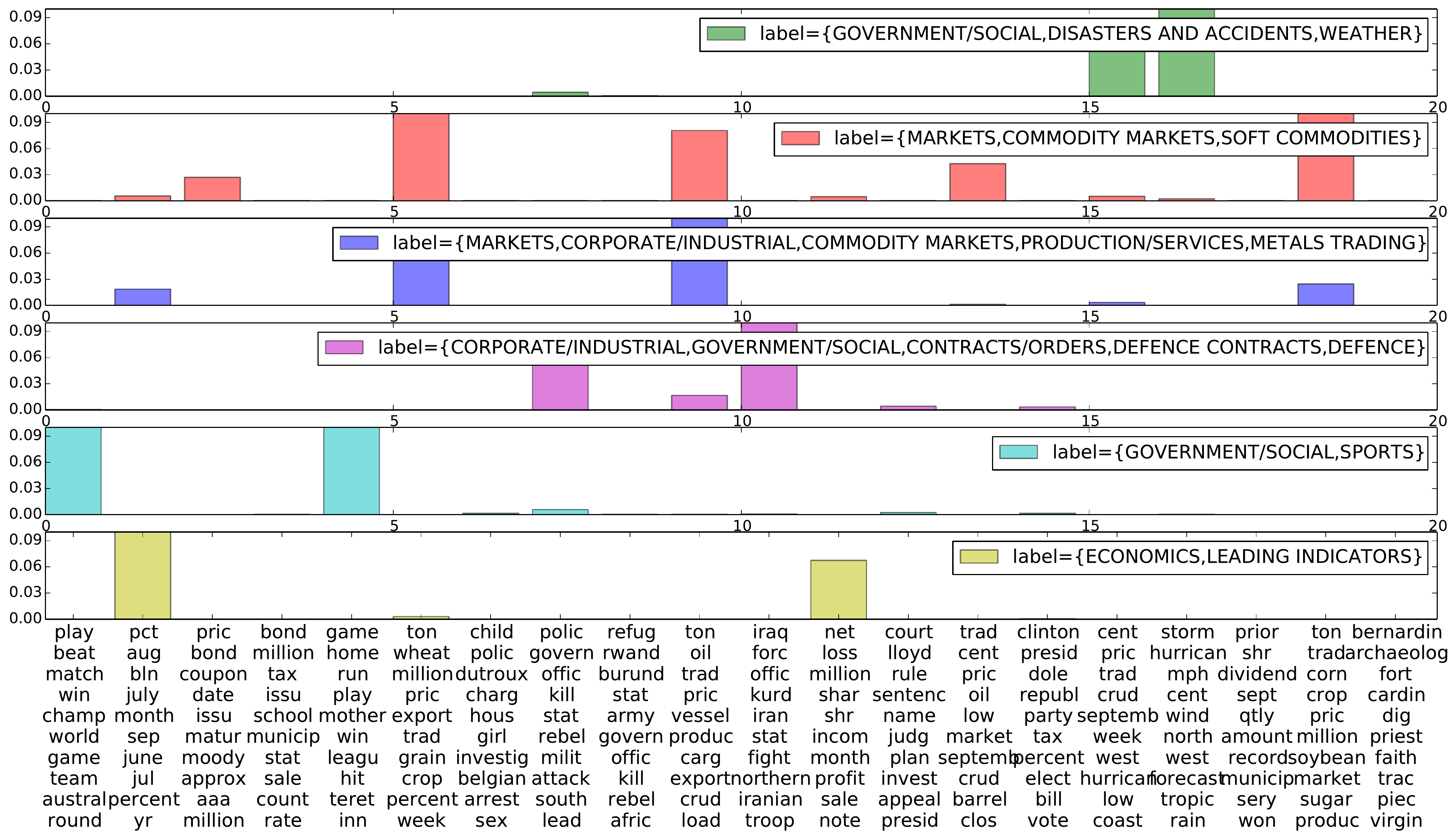}
	}
	\newline
	\subfigure[OHSUMED]{
	\includegraphics[width=0.96\linewidth, clip=true]{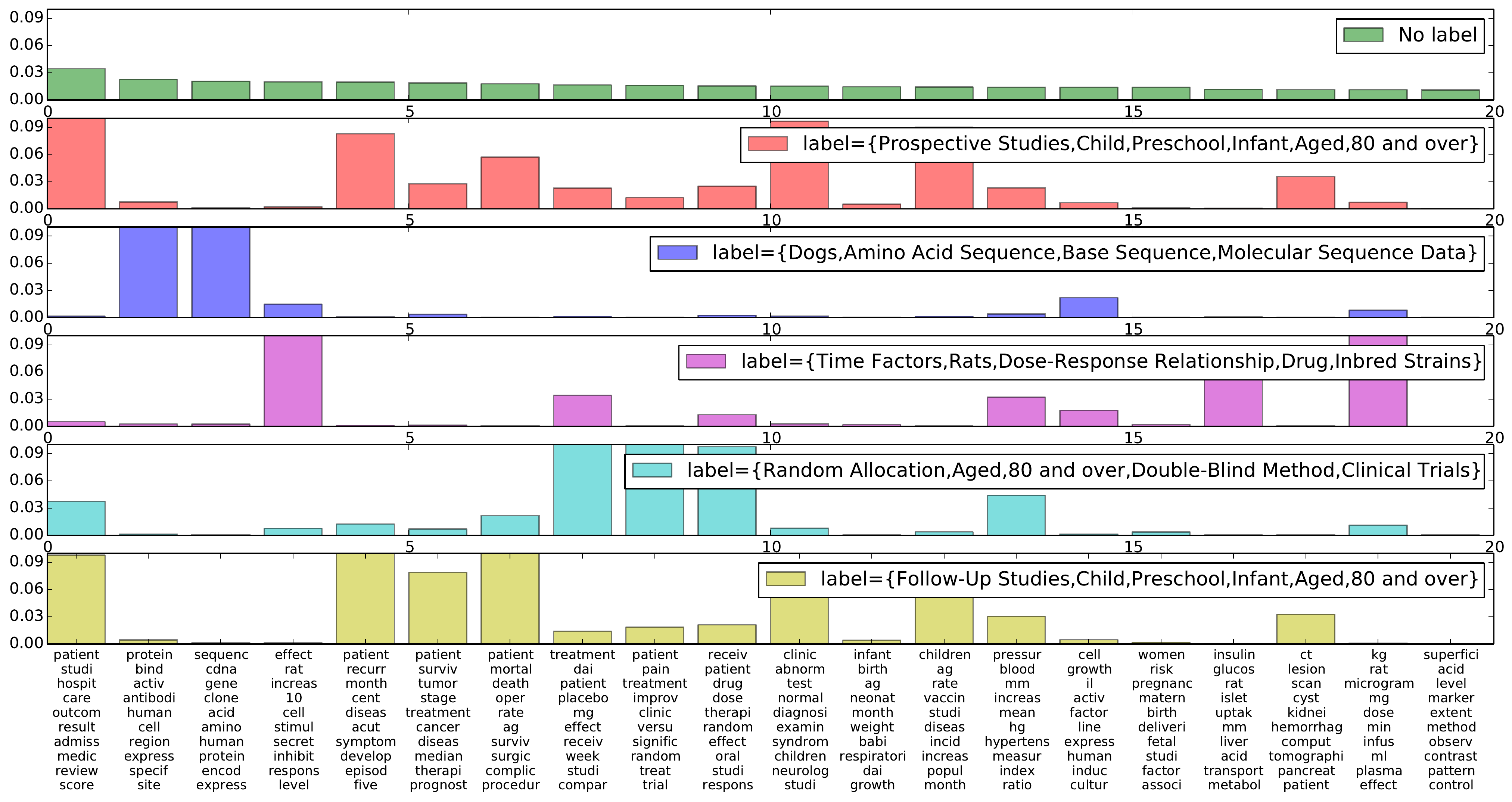}	
	}

	\caption{\label{fig:rcv_topic_label} Expected topic distributions given labels from RCV and OHSUMED. Topics are sorted by their posterior word counts, and the top 20 topics are displayed with the top 10 words (stemmed). From top to bottom, we compute an expected topic distribution given a randomly selected set of labels. }
\end{figure*}

\begin{figure*}[h!]
	\centering
	\includegraphics[width=0.96\linewidth, clip=true]{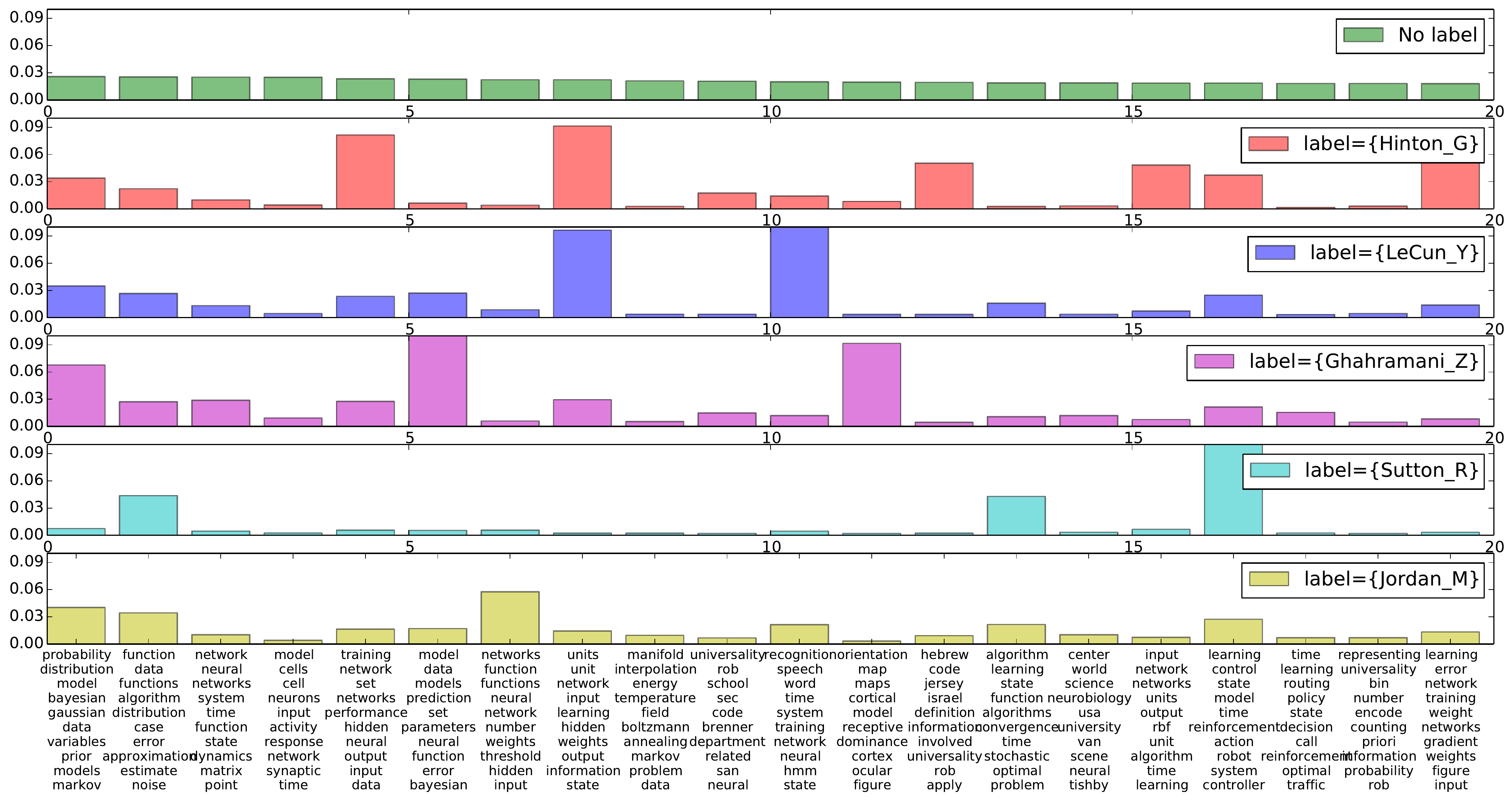}	

	
	\caption{\label{fig:nips_topic_label} Expected topic distributions given labels from NIPS. Topics are sorted by their posterior word counts, and the top 20 topics are displayed with the top 10 words. From top to bottom, we compute an expected topic distribution given a set of representative labels. }
\end{figure*}

Figure \ref{fig:rcv_topic_label}  and \ref{fig:nips_topic_label} show the expected topic distributions of HDSP given different sets of labels. 
When multiple labels are given, the model expects high probabilities for the topics that are similar to all given labels. For example, when `Market' and `Sports' labels are given, the model expects high probabilities on sports related topics and relatively high probability on `Market' related topics based on the weights between topics and two labels.


\subsubsection{Modeling data with missing labels}
We also test our model with partially labeled data which have not been previously covered in topic modeling. Many real-world data fall into this category where some of the data are labeled, others are incompletely labeled, and the rest are unlabeled. For this experiment, we randomly remove existing labels from the RCV and OHSUMED corpora. To remove observed labels in the training corpus, we use Bernoulli trials with varying parameters to analyze how the proportion of observed labels affects the heldout predictive performance of the model.


\begin{figure*}[t!]
	\centering
	\subfigure[OHSUMED\label{fig:ll_ohsumed_partial}]{
	\includegraphics[width=0.45\linewidth, clip=true]{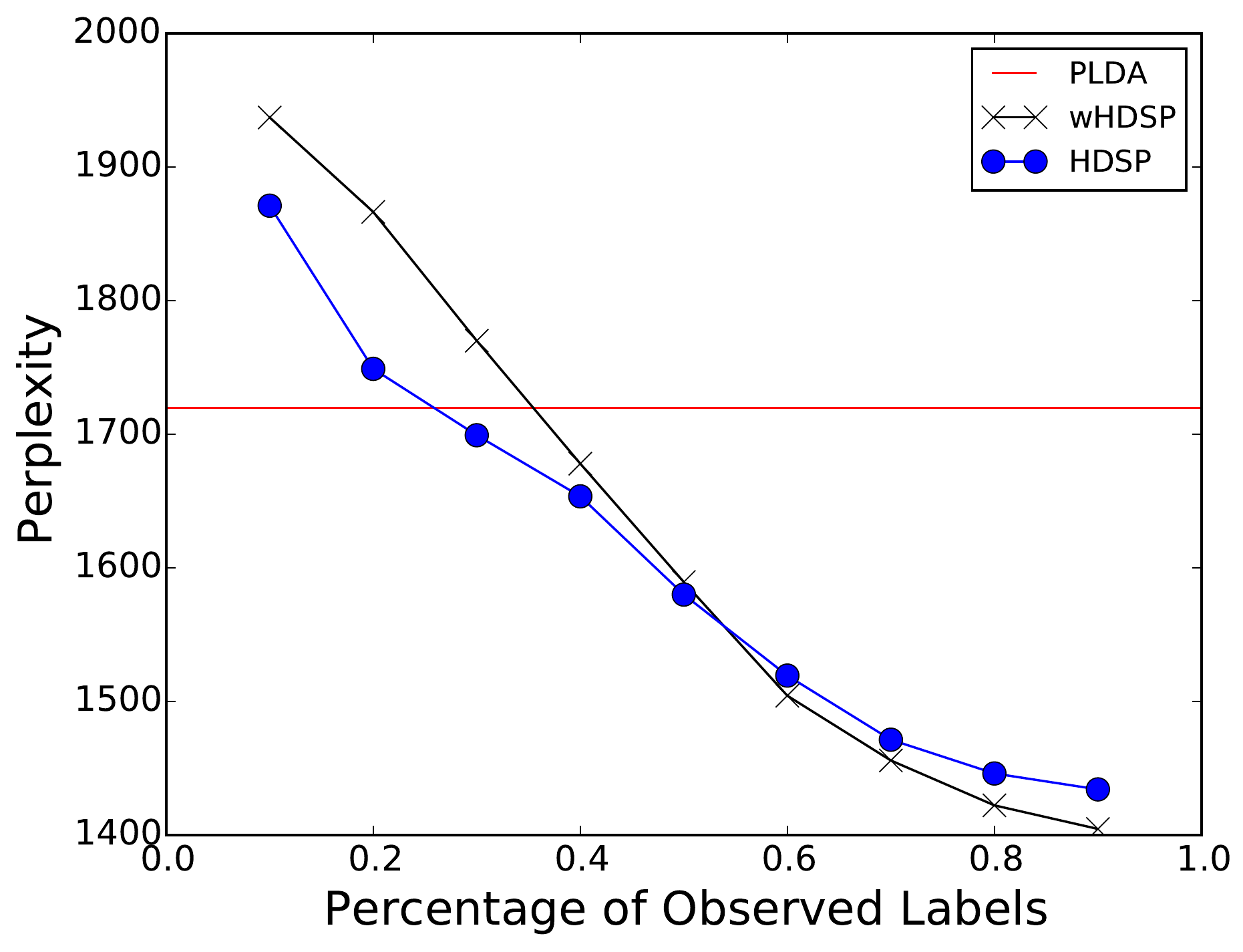}			
	}
	\subfigure[RCV\label{fig:ll_rcv_partial}]{	
	\includegraphics[width=0.45\linewidth, clip=true]{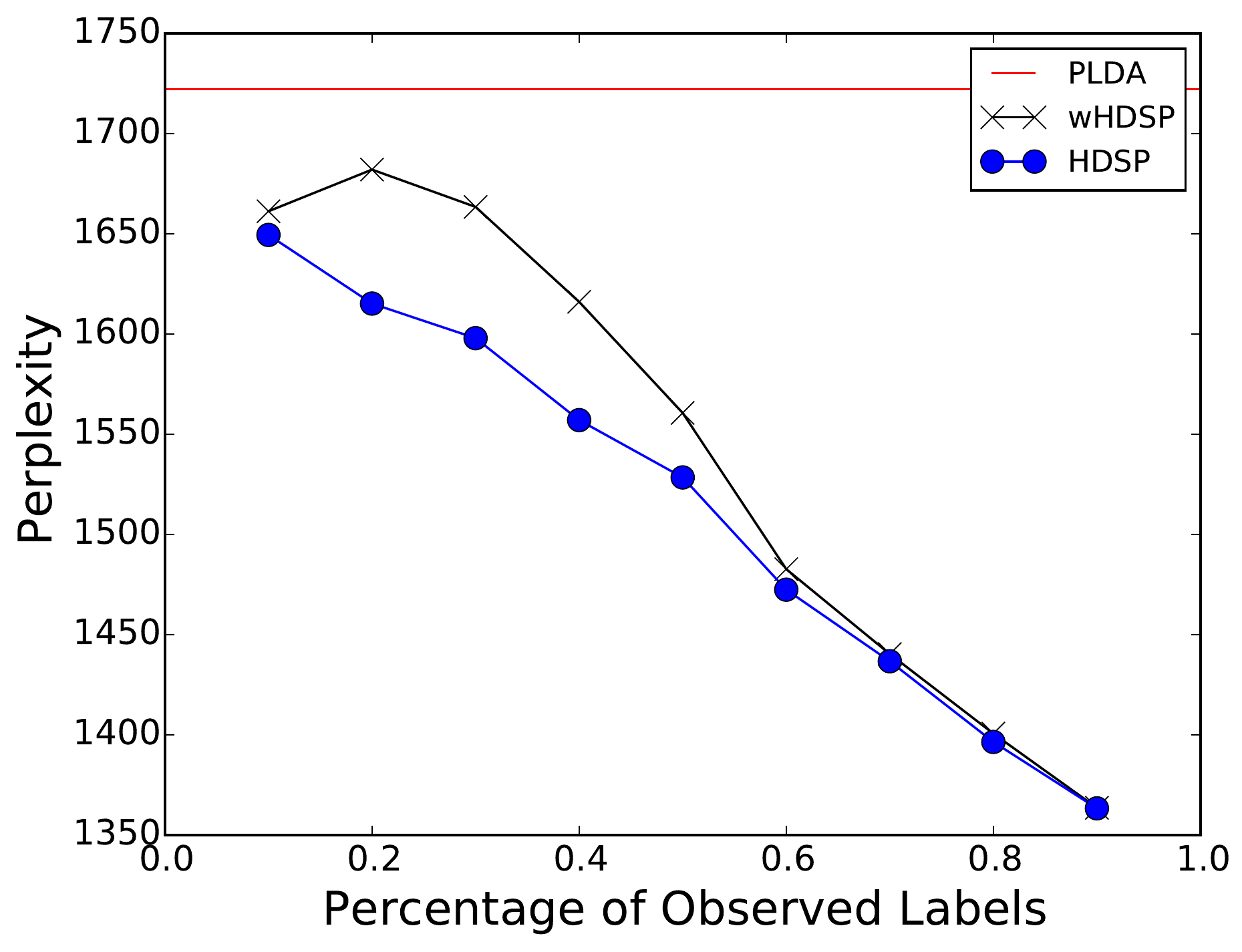}
	}
	\caption{\label{fig:heldout_partial_ll} Perplexity of partially labeled documents. For both RCV and OHSUMED, we randomly remove the observed labels of training documents based on Bernoulli trials. After training the model with removed dataset, we measure the heldout perplexity on a test documents with different scaling functions. }
\end{figure*}

Figure \ref{fig:heldout_partial_ll} shows the predictive perplexity with varying parameters of Bernoulli distribution from 0.1 to 0.9. For both scaling functions, the perplexity decreases as the model observes more labels. Compared to the PLDA (with the parameter setting for optimal performance), the HDSP achieves similar perplexity with only 20\% of the labels. One notable phenomenon is that the HDSP outperforms wHDSP on both datasets when the number of observed labels is less than 50\% of the total number of labels. 

\subsection{Mixed-Type data}
In this section, we present the performance of the second scaling function with a corpus of product reviews which has real-valued ratings and category information. 

\begin{table}[t!]
  \centering
  \caption{\label{amazon_data}The number of reviews for each rating and category in the Amazon dataset. The dataset contains 24,259 reviews collected from seven different product categories. The description reveals a highly skewed distribution of reviews where 52\% of reviews are rated as five-star. The percentage denotes the proportion of reviews per each rating.}
    \begin{tabular}{l|rr}
    &\# reviews & percentage\\\hline
    Total & 24,259 & 100\% \\
    5-star &   12,382 &52\% \\
    4-star & 5,040& 20\% \\
    3-star & 1,905& 8\% \\
    2-star & 1,723& 7\% \\
    1-star & 3,209& 13\%
    \end{tabular}
    \quad{}
    \begin{tabular}{l|r}
    Category & \# reviews \\ \hline
    Canister vacuum & 3535\\
    Digital SLR & 4189\\
    Laptop & 4252\\
    MP3 &   3659\\
    Air conditioner &  568\\
    Space heater &   3859\\
    Coffee machine &  4197\\
    \end{tabular}    
\end{table}

The first scaling function is only applicable to categorical side information, so we use the second scaling function (wHDSP) which can model numerical as well as categorical side information of documents. To evaluate the performance of wHDSP with numerical side information, we train the model with the Amazon review data collected from seven categories of electronic products: air conditioner, canister vacuum, coffee machine, digital SLR, laptop, MP3 player, and space heater. Amazon uses a five-star rating system, so each review contains one numerical rating ranging from one to five. Table \ref{amazon_data} shows the number of reviews for each rating and category.  Recall that $r$ is a vector whose values denote the observation of the labels. For each review, we set the dimension of $r$ to eight in which the first dimension is a numerical rating of a review, and then the remaining seven dimensions match the seven product categories. We set the value of each dimension to one if the review belongs to the corresponding category, and zero otherwise.

\begin{table}[t!]
  \centering
  \small
  \caption{\label{mm_f1pr}F1 of wHDSP and the other models for the Amazon review corpus. wHDSP and SLDA perform comparably on one-star ratings but wHDSP outperforms SLDA on middle range ratings (two, three, and four stars).  }
    \begin{tabular}{l|rrrrr}
          & \multicolumn{5}{c}{Ratings}            \\
    F1    & 1     & 2     & 3     & 4     & 5 \\ \hline
    wHDSP & 0.600 & \textbf{0.161} & \textbf{0.185} & \textbf{0.316} & 0.687 \\
    wHDSP-no-cate & 0.428 & 0.087 & 0.099 & 0.061 & 0.658 \\
    LDA50+SVM & 0.392 & 0.036 & 0.038 & 0.134 & 0.684 \\
    LDA100+SVM & 0.454 & 0.078 & 0.073 & 0.265 & 0.678 \\
    LDA200+SVM & 0.508 & 0.032 & 0.100 & 0.284 & 0.681 \\
    SLDA50 & 0.603 & 0.000 & 0.021 & 0.140 & \textbf{0.741} \\
    SLDA100 & 0.606 & 0.000 & 0.021 & 0.067 & 0.740 \\
    SLDA200 & 0.580 & 0.015 & 0.011 & 0.140 & 0.727 \\
	SVM&    0.403 & 0.000     & 0.000     & 0.007 & 0.716 \\
	NaiveBayes&    \textbf{0.634} & 0.028 & 0.085 & 0.469 & 0.652 \\
	DecisionTree&    0.457 & 0.088 & 0.154 & 0.355 & 0.628    
    \end{tabular}%
  \label{tab:addlabel}%
\end{table}%


\begin{table}[t!]
  \centering
  \caption{\label{mm_whdsp}Macro and micro F1 of the wHDSP and the other models. The left table shows the performance of classification with five-star ratings system. The models with asterisk are trained without category information. Note that the SLDA cannot incorporate two different types of labels together.}
    \begin{tabular}{l|rr}
    5-Ratings & {MacroF1} & MicroF1 \\ \hline
    wHDSP & \textbf{0.390} & 0.522 \\
    wHDSP* & 0.267 & 0.474 \\
    LDA50+SVM & 0.257 & 0.518 \\
    LDA100+SVM & 0.310 & 0.520 \\
    LDA200+SVM & 0.321 & 0.527 \\
    LDA200+SVM* & 0.309 & 0.533 \\
    SLDA50 & 0.301 & 0.584 \\
    SLDA100 & 0.287 & \textbf{0.588} \\
    SLDA200 & 0.294 & 0.577 \\
	SVM & 0.225 & 0.560\\
	NaiveBayes & 0.374 & 0.545\\
	DecisionTree&0.336&0.477    
    \end{tabular}%
\end{table}%

To evaluate the performance of wHDSP, we classify the ratings of the reviews based on a trained model. We use 90\% of the corpus to train models and the remaining 10\% of the corpus to test the models. To classify the rating of each review in the test set, we compute the perplexity of the given review with varying ratings from one to five, and choose the rating that shows the lowest perplexity. Generally, computing the perplexity of heldout document requires complex approximation schemes \citep{Wallach:2009p5474}, but we compute the perplexity based on the expected topic distribution given category and rating information, which requires a finite number of computations.

We compare the wHDSP with the supervised LDA (SLDA), LDA$+$SVM, as well as classifiers Naive Bayes, SVM, and decision trees (CART). For the LDA$+$SVM approach, we first train the LDA model and then use the inferred topic proportion and categories as features of the SVM. For the SLDA model, the category information cannot be used because the model is designed to learn and predict the single response variable. For both models, we set the number of topics to 50, 100, and 200.

\begin{figure}[t!]
	\centering
	
	\begin{tabular}{lll}
	\subfigure[wHDSP]{
	\includegraphics[width=0.3\linewidth]{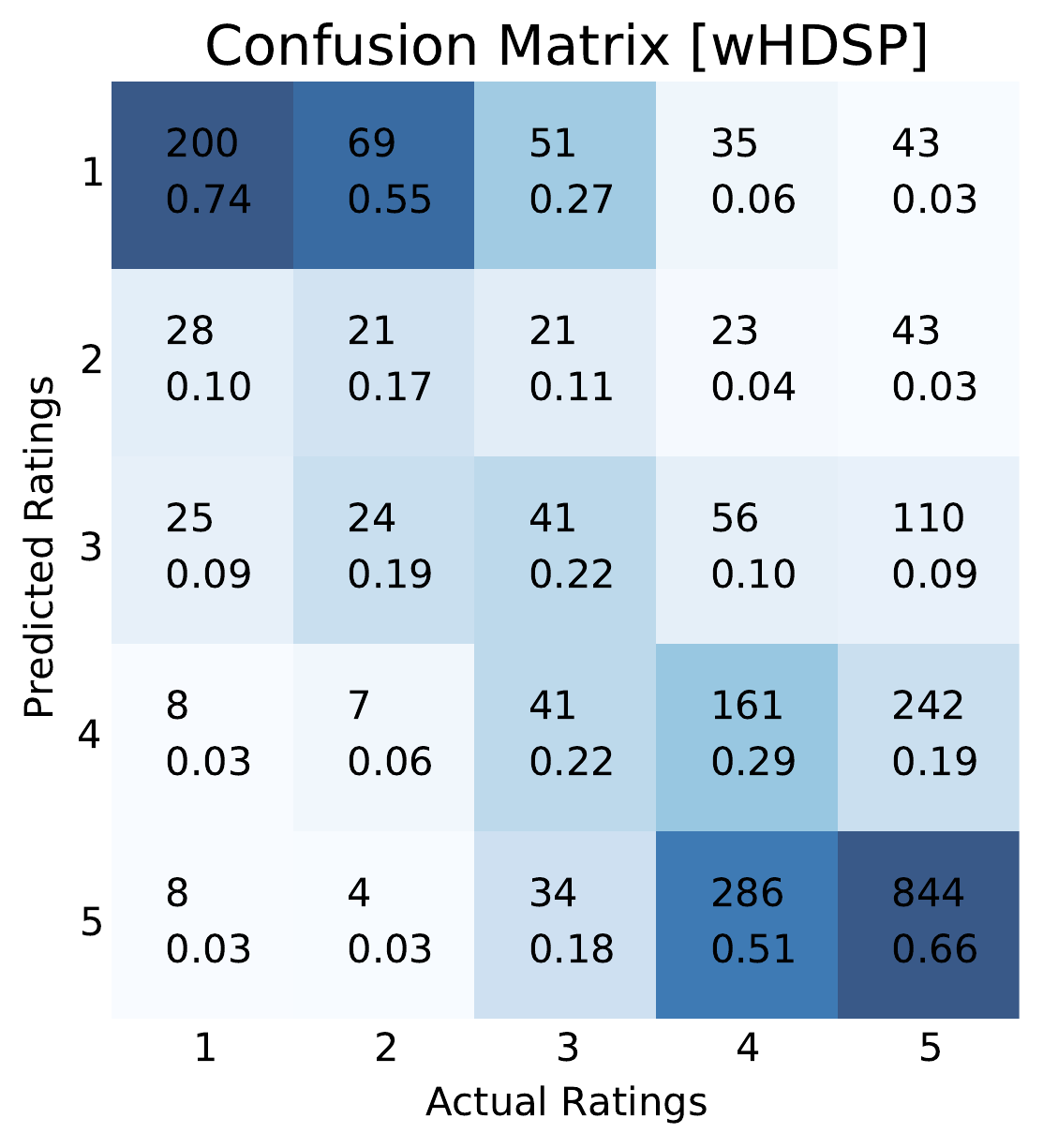}
	}&
	\subfigure[\label{fig:conf_without}wHDSP without categories]{
	\includegraphics[width=0.3\linewidth]{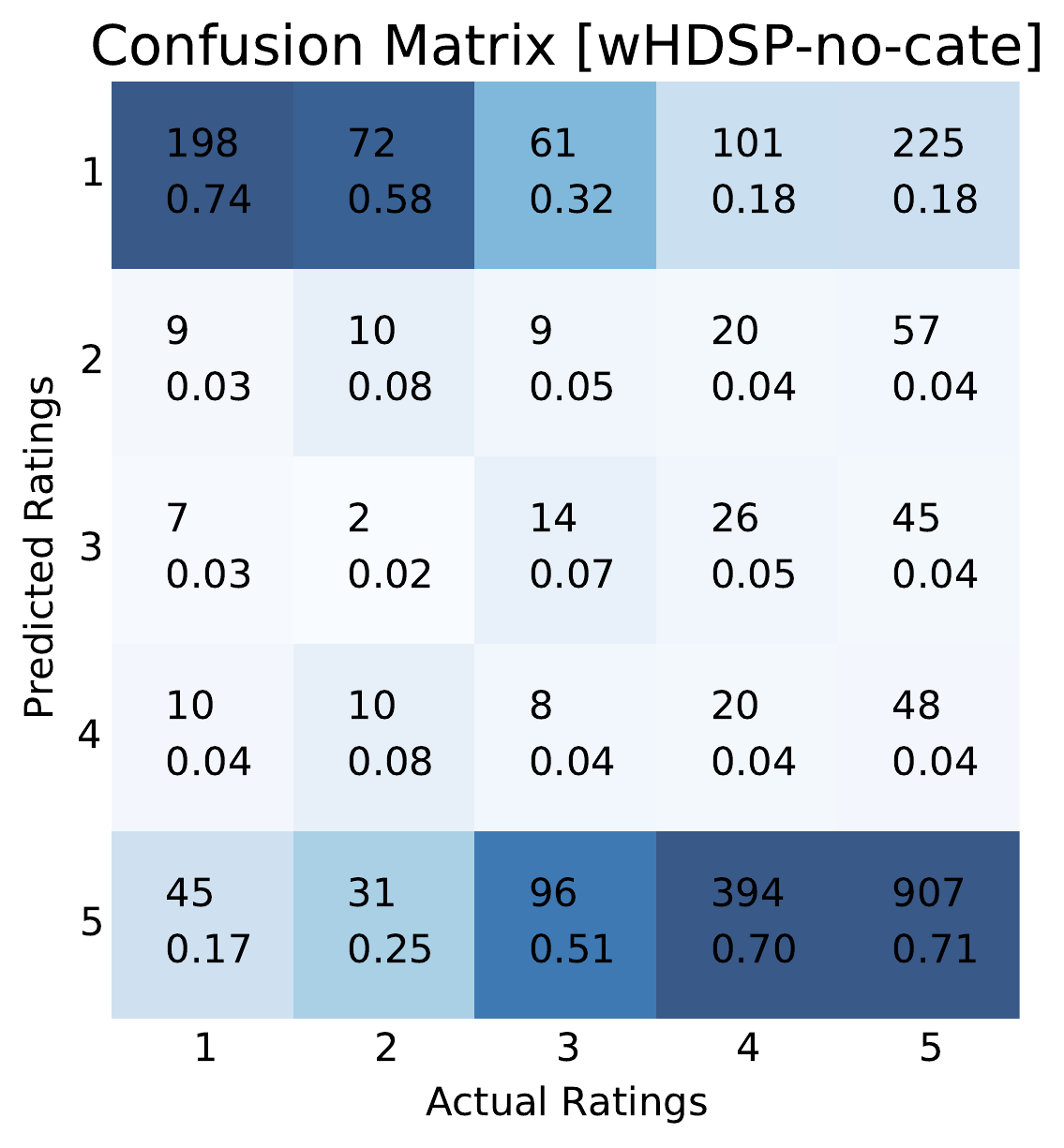}	
	}&
	\subfigure[\# of instances per rating]{
	\includegraphics[width=0.3\linewidth]{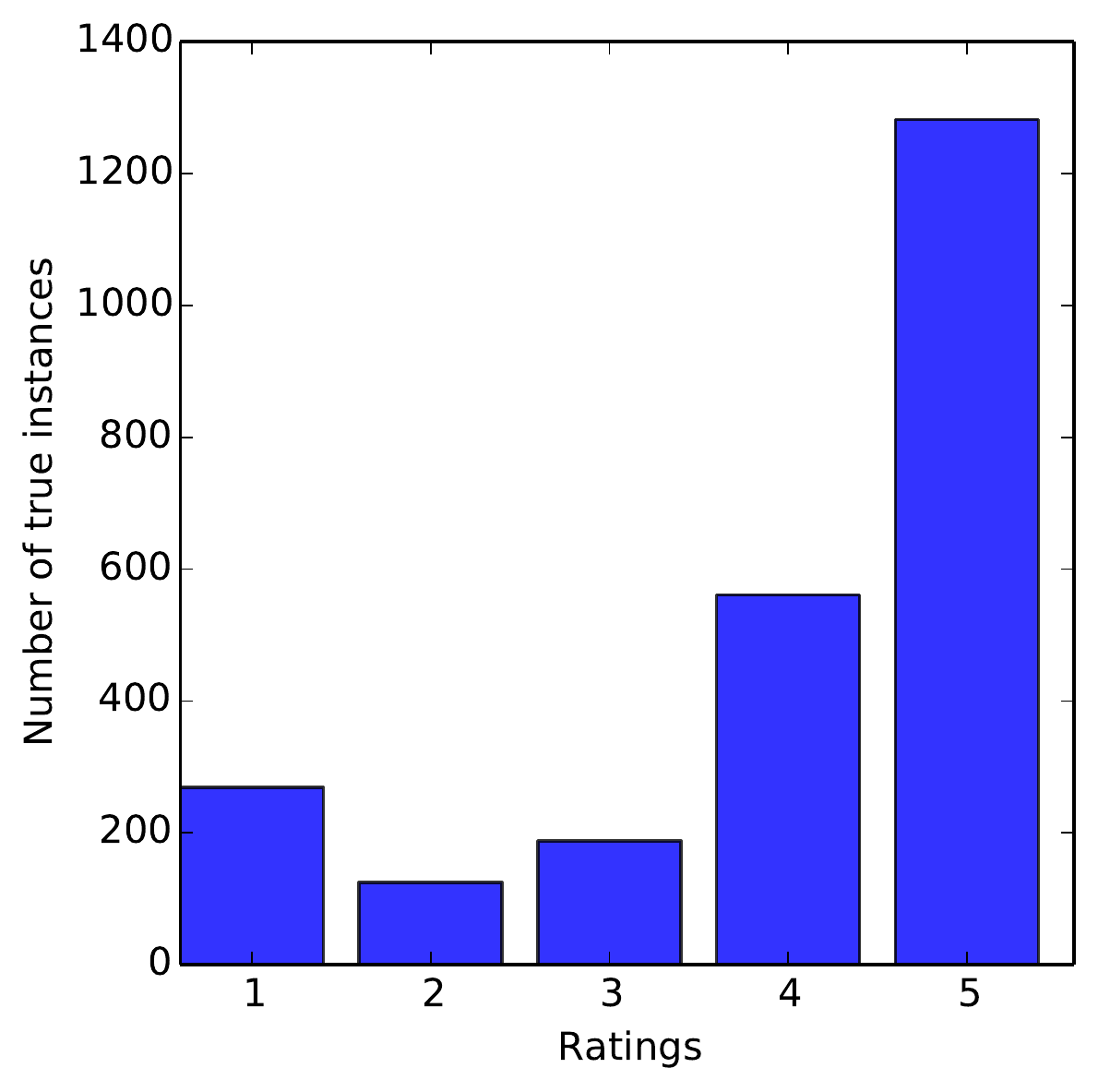}	
	}\\

	\subfigure[SLDA50]{
	\includegraphics[width=0.3\linewidth]{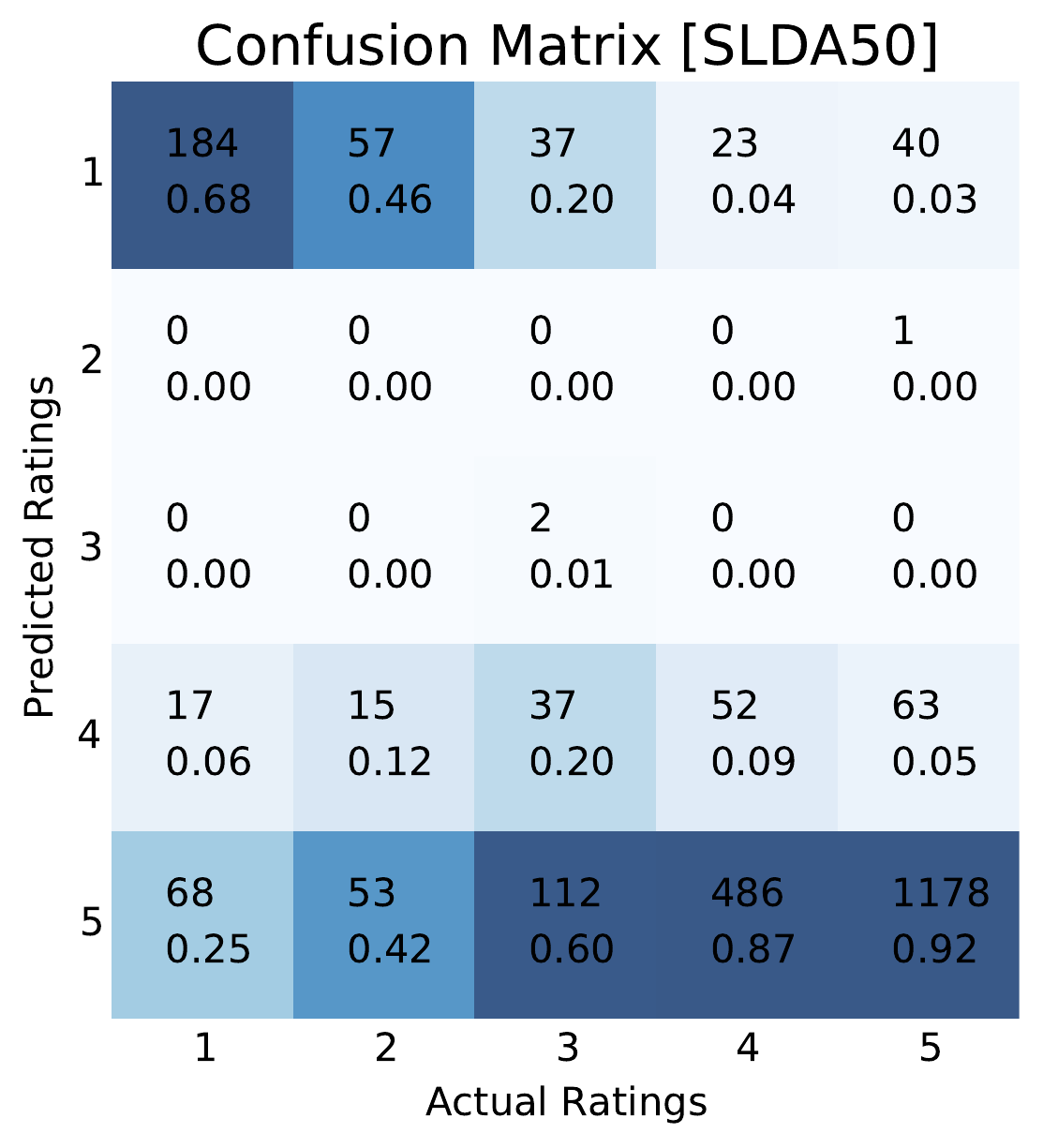}
	}&
	\subfigure[LDA200+SVM]{
	\includegraphics[width=0.3\linewidth]{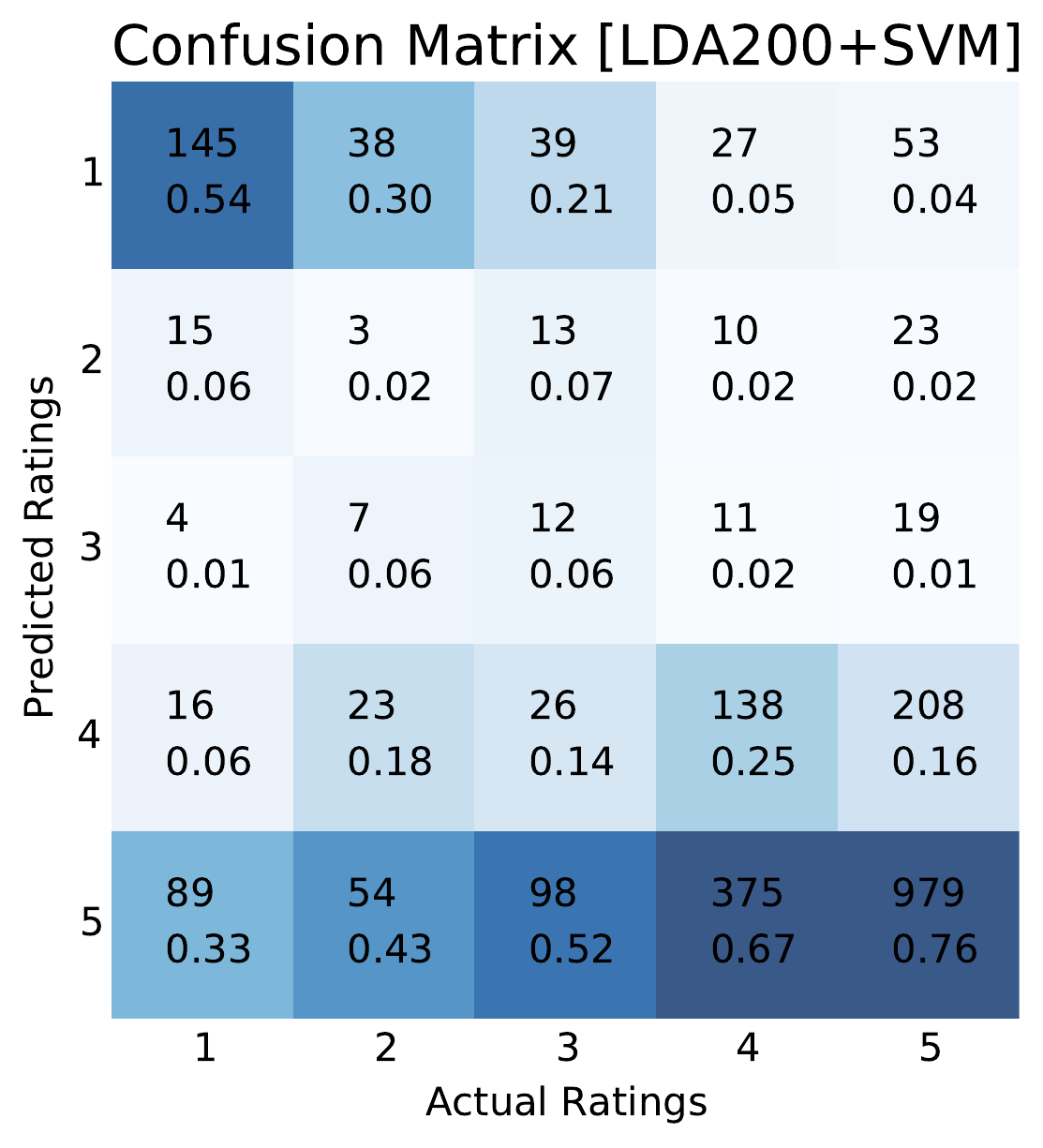}	
	}& 	
	\subfigure[SVM]{
	\includegraphics[width=0.3\linewidth]{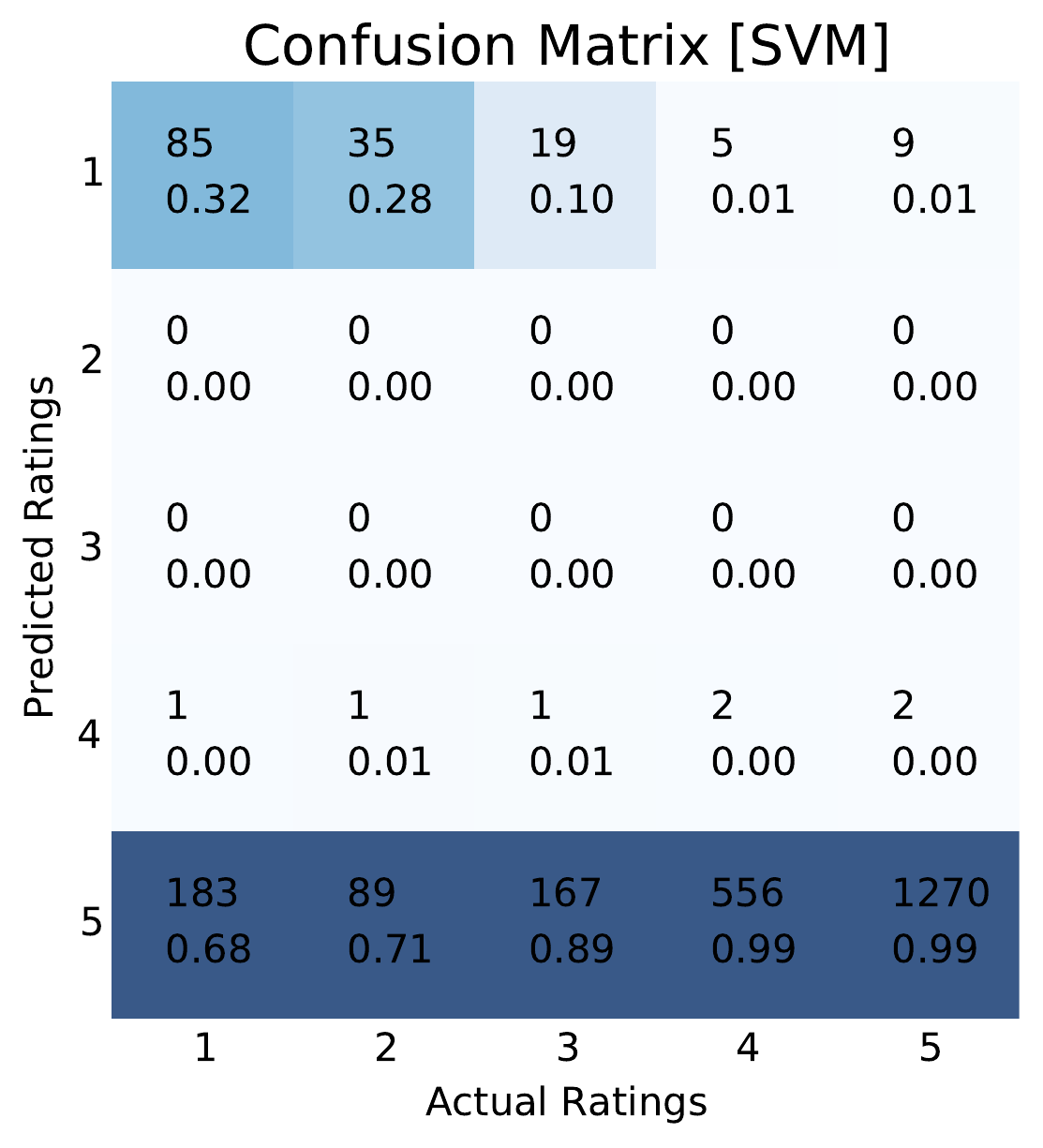}	
	}\\	
	\subfigure[NaiveBayes]{
	\includegraphics[width=0.3\linewidth]{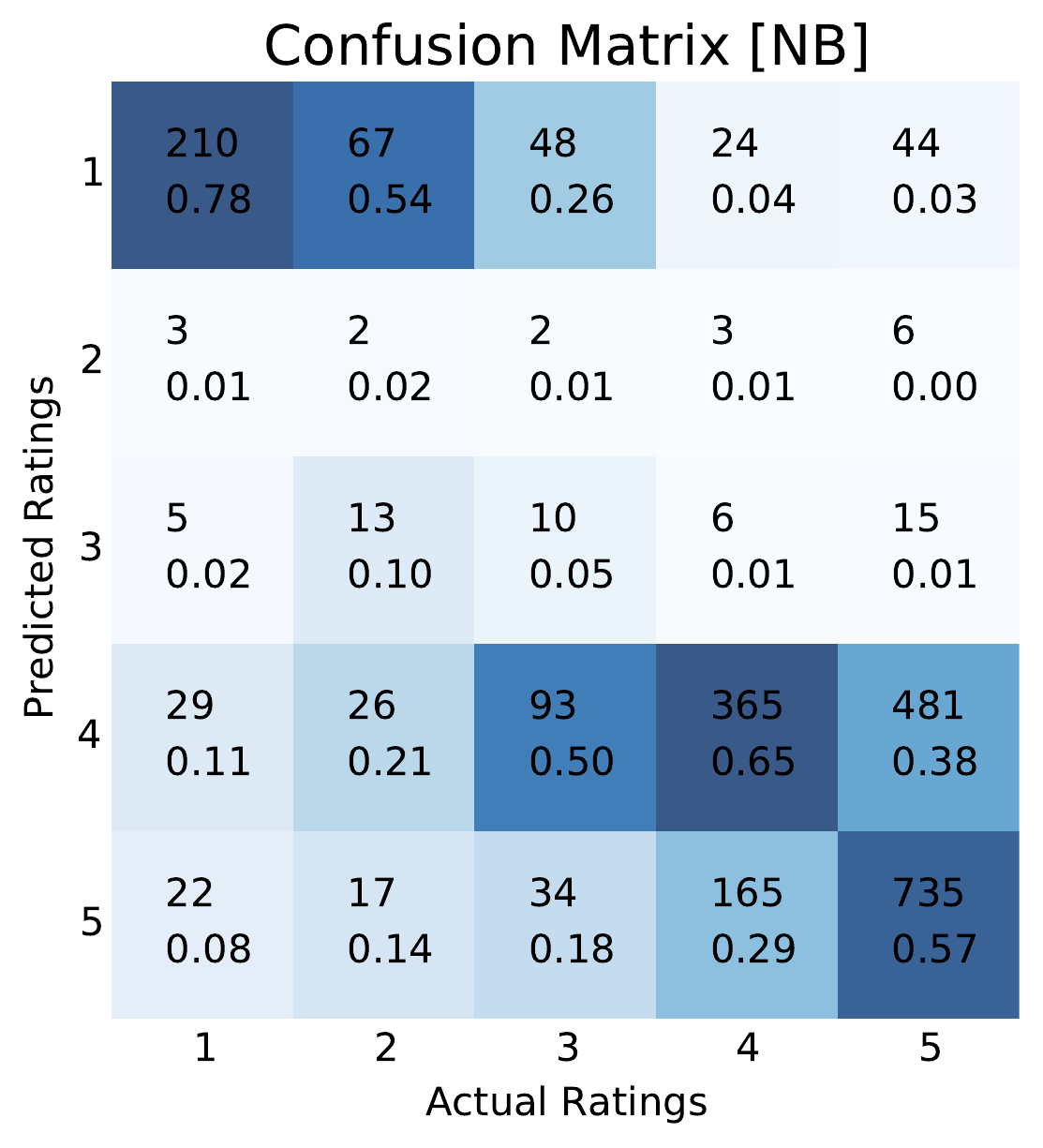}	
	}&
	
	\subfigure[DecisionTree]{
	\includegraphics[width=0.3\linewidth]{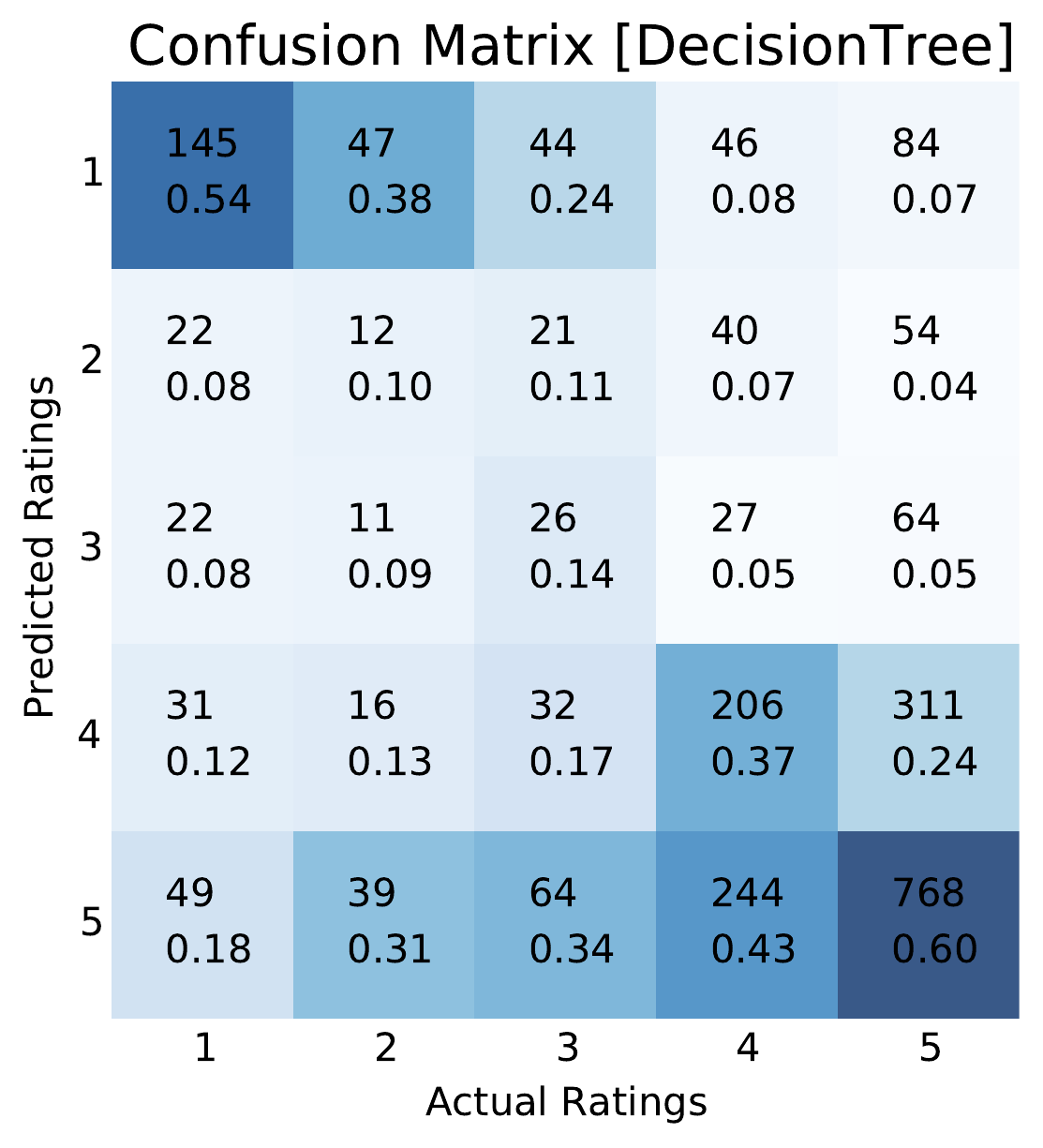}	
	}
	
	\end{tabular}
	\caption{\label{fig:conf_matrix} Confusion matrices from classification results of wHDSP and the other models. Diagonal entries indicate the number of correctly classified reviews per rating. Despite the highly skewed data distribution (c), wHDSP achieves relatively better classification results for negative and neutral reviews as shown in (a). Confusion matrices are column-wise normalized.}
\end{figure}

In many applications, classifying negative feedback of users is more important than classifying positive feedback. From the negative feedback, companies can identify possible problems of their products and services and use the information to design their next product or improve their services. In most online reviews, however, the proportion of negative feedback is smaller than the proportion of positive feedback. For example, in the Amazon data, about 51\% of reviews are rated as five-star, and 72\% rated as four or five. A classifier trained by such skewed data is likely to be biased toward the majority class.

We report the classification results of Amazon dataset in Table \ref{mm_f1pr} and Table \ref{mm_whdsp}. Table \ref{mm_f1pr} shows the results for each rating in terms of F1, and Table \ref{mm_whdsp} shows the results in terms of micro and macro F1. The wHDSP outperforms the other models in terms of macro F1 but performs worse than sLDA in terms of micro F1. As we noted earlier, classifying negative reviews may be more important in many applications. Both the SLDA with 100 topics and the wHDSP are comparable in  classifying the most negative (one-star) reviews. However, the confusion matrices and Table \ref{mm_f1pr} indicate that the SLDA dichotomously learns the decision boundaries where the most reviews are classified into one-star or five-star. For example, the SLDA with 50 and 100 topics did not classify any two-star and three-star reviews correctly. The wHDSP learns the decision boundaries for classifying subtle differences between five-star rating reviews. These patterns are shown clearly with the confusion matrices in Figure \ref{fig:conf_matrix} where the diagonal entries are the numbers of correctly classified reviews. For example, the wHDSP classified only eight one-star reviews as five-star reviews, but the SLDA50 assigned 68 reviews as five-star reviews. 

We perform a rating prediction task with and without the category information of reviews to see the effect of using both the category and rating information on the wHDSP and LDA+SVM approaches. The results represented by wHDSP* in Table \ref{mm_whdsp} and Figure \ref{fig:conf_without} show the performances of rating prediction with the wHDSP trained without category information. For wHDSP*, the model performs worse than wHDSP, which indicates the model, without category information, cannot distinguish the review ratings which depend on topical context. The LDA$+$SVM without categories achieves 0.309 macro F1 and 0.533 micro F1, which are comparable to the LDA+SVM with the category information. Unlike the wHDSP, the decision boundaries of SVM are not improved with the additional category information. The result supports that for learning decision boundaries between ratings over different categories, the approach of including category information to train topics is more effective than using topics and the category information independently.

\section{\label{sec:con}Discussions}
We have presented the hierarchical Dirichlet scaling process (HDSP), a Bayesian nonparametric prior for a mixed membership model that lets us analyze underlying semantics and observable side information. The combination of the stick breaking process with the normalized gamma process in HDSP is a more controllable construction of the hierarchical Dirichlet process because each atom of the second level measure inherits from the first level measure in order. HDSP also allows more flexibility and the capability of modeling side information by the scaling functions that plug into the rate parameter of the gamma distribution. The choice of the scaling function is the most important part of the model in terms of establishing a link between topics and observed labels. We developed two scaling functions but the choice of scaling function depends on the modeler's intention. For example, the well known linking functions from the generalized linear model can be used as scaling functions, or one can use several scaling functions together on purpose. We showed that the application of HDSP to topic modeling correctly recovers the topics and topic-label weights of synthetic data.  Experiments with the real dataset show that the first scaling function is more suited for partially labeled data, and the second first scaling function is more suited for a dataset with both numerical and categorical labels.

Hierarchical Dirichlet scaling process opens up a number of interesting research questions that should be addressed in future work. First, in the two scaling functions we proposed to model the correlation structure between topics and side information, we simply defined the relationship between topic $k$ and label $j$ through the scaling parameter $w_{kj}$. However, this approach does not consider the correlation within topics and labels. Taking inspiration from previous work \citep{blei2007correlated, mimno2007mixtures, paisley2012discrete} that showed correlations among topics, we can define a scaling function with a prior over the topics and labels to capture their complex relationships. Second, our posterior inference algorithm based on mean-field variational inference is tested with tens of thousands documents. However, modern data analysis requires inference of massive and/or streaming data. For a fast and efficient posterior inference, we can apply parallel or distributed algorithms based on a stochastic update \citep{hoffman2013stochastic, ahn2014distributed}. Furthermore, we fix the number of labels before training but we need to find a way to model the unbounded number of labels for  streaming data.




\appendix

\section*{\label{sec:vi_detail}Appendix A. Variational inference for HDSP}
In this section, we provide the detailed derivation for mean-field variational inference for HDSP with the first scaling function. First, the evidence of lower bound for HDSP is obtained by taking a Jensen's inequality on the marginal log likelihood of the observed data,
\begin{align}
\ln \int p(\mathcal{D}, \Theta) d\Theta \geq \int Q(\Psi) \ln \frac{P(\mathcal{D}, \Theta)}{Q(\Psi)}  d\Theta,
\end{align}
where $\mathcal{D}$ is the training set of documents and labels. $\Psi$ denotes the set of variational parameters, $\Theta$ denotes the set of model parameters.

We define a fully factorized variational distribution $Q$ as follows:
\begin{align}
Q := \prod_{k=1}^{T} q(\phi_k)q(V_k) \prod_{j=1}^J q(w_{kj}) \prod_{m=1}^{M} q(\pi_{mk}) \prod_{n=1}^{N_m} q(z_{mn}),
\end{align}
where
\begin{align}
&q(z_{mn}) = \text{Multinomial}(z_{mn} | \gamma_{mn1}, \gamma_{mn2}, ..., \gamma_{mnT})&\\
&q(\pi_{mk}) = \text{Gamma}(\pi_{mk} | a^\pi_{mk}, b^\pi_{mk})&\notag\\
&q(w_{kj}) = \text{InvGamma}(w_{kj} | a^w_{kj}, b^w_{kj})&\notag\\
&q(\phi_k) = \text{Dirichlet}(\phi_k | \eta_{k1}, \eta_{k2}, ..., \eta_{kI})&\notag\\
&q(V_k) = \delta_{V_k}.&\notag
\end{align}
The evidence of lower bound (ELBO) is
\begin{align}
&L(\mathcal{D}, \Psi) = \mathbb{E}_q[\ln p(\mathcal{D}, \Psi)] + \mathbb{H}[Q] &\\
&= \mathbb{E}_q[ \sum_{m=1}^{M} \sum_{n=1}^{N_m}\ln p(\mathrm{x}_{mn}|z_{mn}, \Phi)] + \mathbb{E}_q[\sum_{m=1}^{M} \sum_{n=1}^{N_m} \ln p(z_{mn}|\pi_{m})] &\notag\\
&\quad +\mathbb{E}_q[\sum_{m=1}^{M} \sum_{k=1}^{\infty} \ln p(\pi_{mk}|V_k, w_k, \mathbf{r}_m)] + \mathbb{E}_q[\sum_{k=1}^{\infty}\ln p(V_k|\alpha)] + \mathbb{E}_q[\sum_{j=1}^{J}\sum_{k=1}^{\infty} \ln p(w_{kj}|a^w,b^w)] &\notag\\
&\quad + \mathbb{E}_q[\sum_{k=1}^{\infty} \ln p(\phi_k|\eta)]  - \mathbb{E}_q[\ln Q]  &\notag\\
&= \sum_{m=1}^{M} \sum_{n=1}^{N_m} \sum_{k=1}^{T} \gamma_{mnk} \mathbb{E}_q[ \ln p(\mathrm{x}_{mn}|\phi_k) ] +\sum_{m=1}^{M}\sum_{n=1}^{N} \sum_{k=1}^{T} \gamma_{mnk} \mathbb{E}_q[\ln  p(z_{mn}=k|\pi_m) ] &\notag\\
&\quad +\sum_{m=1}^{M} \sum_{k=1}^{T} \mathbb{E}_q[\ln p(\pi_{mk}|V_k, w_k, \mathbf{r}_m)] + \sum_{k=1}^{T} \mathbb{E}_q[\ln p(V_k|\alpha)] + \sum_{k=1}^{\infty} \sum_{j=1}^{J} \mathbb{E}_q[\ln p(w_{kj}|a^w, b^w)]   - \mathbb{E}_q[\ln Q] &\notag
\end{align}
where the expectations of latent variables under the variational distribution $Q$ are 
\begin{align}
&\mathbb{E}_q[\pi_{mk}] = a^\pi_{mk}/b^\pi_{mk} & \notag\\
&\mathbb{E}_q[\ln \pi_{mk}] = \psi(a^\pi_{mk}) - \ln b^\pi_{mk} &\notag\\
&\mathbb{E}_q[w_{kj}] = b^w_{kj}/(a^w_{kj}-1)&\notag\\
&\mathbb{E}_q[w_{kj}^{-1}] = a^w_{kj}/b^w_{kj}&\notag\\
&\mathbb{E}_q[\ln w_{kj}] = \ln b^w_{kj} - \psi(a^w_{kj})&\notag\\
&\mathbb{E}_q[\ln \phi_{ki}] = \psi(\eta_{kd}) - \psi(\sum_i' \psi_{ki'})&\notag
\end{align}
Then, we derive the equations further
\begin{align}
&L(\mathcal{D}, \Psi)= \sum_{m=1}^{M} \sum_{n=1}^{N_m} \sum_{k=1}^{T} \gamma_{mnk} \{\psi(\eta_{k\mathrm{x}_{mn}}) - \psi(\sum_d \eta_{kd})\} &\\
&\quad + \sum_{m=1}^{M} \sum_{n=1}^{N_m} \sum_{k=1}^{T} \gamma_{mnk} \{ \mathbb{E}_q[\ln \pi_{mk}] - \mathbb{E}_q[\ln\sum_{k=1}^T \pi_{mk}] \}  &\notag\\
&\quad + \sum_{m=1}^{M} \sum_{k=1}^{T} -\beta p_k \sum_j r_{mj} \{\ln (b^w_{kj}) - \psi(a^w_{kj}) \} + (\beta p_k -1)\{\psi(a^\pi_{mk}) - \ln(b^\pi_{mk}) \} & \notag\\
&\quad\quad - \prod_j \left(\frac{a^w_{kj}}{b^w_{kj}}\right)^{r_{mj}}\frac{a^\pi_{mk}}{b^\pi_{mk}} - \ln \Gamma(\beta p_k) &\notag\\
&\quad + \sum_{k=1}^{T} \ln \Gamma (\alpha + 1) - \ln \Gamma(\alpha) + (\alpha-1)\ln(1-V_k)&\notag\\
&\quad + \sum_{k=1}^{T}\sum_{j=1}^{J} a^w \ln b^w - \ln \Gamma(a^w) - (a^w+1)\{\ln b^w - \psi(a^w)\} - a^w  - \mathbb{E}_q[\ln Q].&\notag
\end{align}
Taking the derivatives of this lower bound with respect to each variational parameter, we can obtain the coordinate ascent updates.

The optimal form of the variational distribution can be obtained by exponentiating the variational lower bound with all expectations except the parameter of interest \citep{bishop2006pattern}. For $\pi_{mk}$, we can derive the optimal form of variational distribution as follows
\begin{align}
&q(\pi_{mk}) \propto \exp \left\{ \mathbb{E}_{q_{-\pi_{mk}}} [\ln p(\pi_{mk}|z, \pi_{-mk}, \mathbf{r}_m, V) ] \right\}	& \\
&\propto \exp \left\{\mathbb{E}_{q_{-\pi_{mk}}} [\ln p(z|\pi_{m}) + \ln p(\pi_{m}|\mathbf{r}_m, V)] \right\} &\notag\\
&\propto \pi_{mk}^{\beta p_k + \sum_{n=1}^{N_m} \gamma_{mnk} - 1 } e^{(b^\pi_{mk} \prod_j \mathbb{E}_q[w_{kj}^{-r_{mj}}] + \frac{N_m}{\xi_m})\pi_{mk}} &\notag
\end{align}
where update for $\xi_m$ is $ - \ln \xi_m - ({\sum_{k=1}^{T}\mathbb{E}_q[\pi_{mk}] - \xi_m})/{\xi_m}$.
Therefore, the optimal form of variational distribution for $\pi_{mk}$ is
\begin{align}
&q(\pi_{mk}) \sim \text{Gamma}(\beta p_k + \sum_{n=1}^{N_m} \gamma_{mnk}, \quad b^\pi_{mk} \prod_j \mathbb{E}_q[w_{kj}^{-r_{mj}}] + \frac{N_m}{\xi_m}).&
\end{align}
We take the same approach described in \citep{paisley2012discrete}, and the only difference comes from the product of the inverse distance term.

For $w_{kj}$, we can derive the optimal form of the variational distribution as follows
\begin{align}
&q(w_{kj}) \propto \exp \left\{ \mathbb{E}_{q_{-w_{kj}}} [\ln p(w_{kj} | \pi, w_{-jk}, a^w, b^w) ] \right\}	& \\
&\propto \exp\left\{ \mathbb{E}_{q_{-w_{kj}}}[ \sum_{m=1}^{M} \ln p(\pi_{mk}|\beta p_k, \prod_{j=1}^J w_{kj}^{-r_{mj}}) + \ln p(w_{kj} | a^w, b^w) ] \right\} &\notag\\
&\propto \exp \left\{ \mathbb{E}_{q_{-w_{kj}}}[ -\beta p_k\sum_{m=1}^{M} r_{mj} \ln w_{kj} - \sum_m \prod_j w_{kj}^{-r_{mj}} \pi_{mk} - (a^w +1) \ln w_{kj} - \frac{b^w}{w_{kj}}  ] \right\}	&\notag\\
&\propto w_{kj}^{-\mathbb{E}_q[\beta {p_k}] \sum_{m=1}^{M} r_{mj} - a^w - 1} e^{(- \sum_{\{m:r_{mj}=1\}} \prod_{\{j':r_{mj'} = 1 /j\}} \mathbb{E}_q[w_{j'k}^{-1}] \mathbb{E}_q[ \pi_{mk}] - b^w )\frac{1}{w_{kj}} }	&\notag
\end{align}
Therefore, the optimal form of variational distribution for $w_{kj}$ is
\begin{align}
&q(w_{kj}) \sim \text{InvGamma}(\mathbb{E}_q[ \beta p_k ] \sum_m r_{mj} + a^w, \quad \sum_{m'}\prod_{j'/j}\mathbb{E}_q[w_{j'k}^{-1}] \mathbb{E}_q[ \pi_{m'k}] + b^w)&
\end{align}
where $m' = \{m:r_{mj}=1\}$ and $j'/j$ $=$ $\{j':r_{mj'} = 1, j' \neq j\}$. 

\section*{B. Posterior Word Count}
\begin{figure}[h]
	\centering
	\subfigure[HDP]{
	\includegraphics[width=0.45\linewidth]{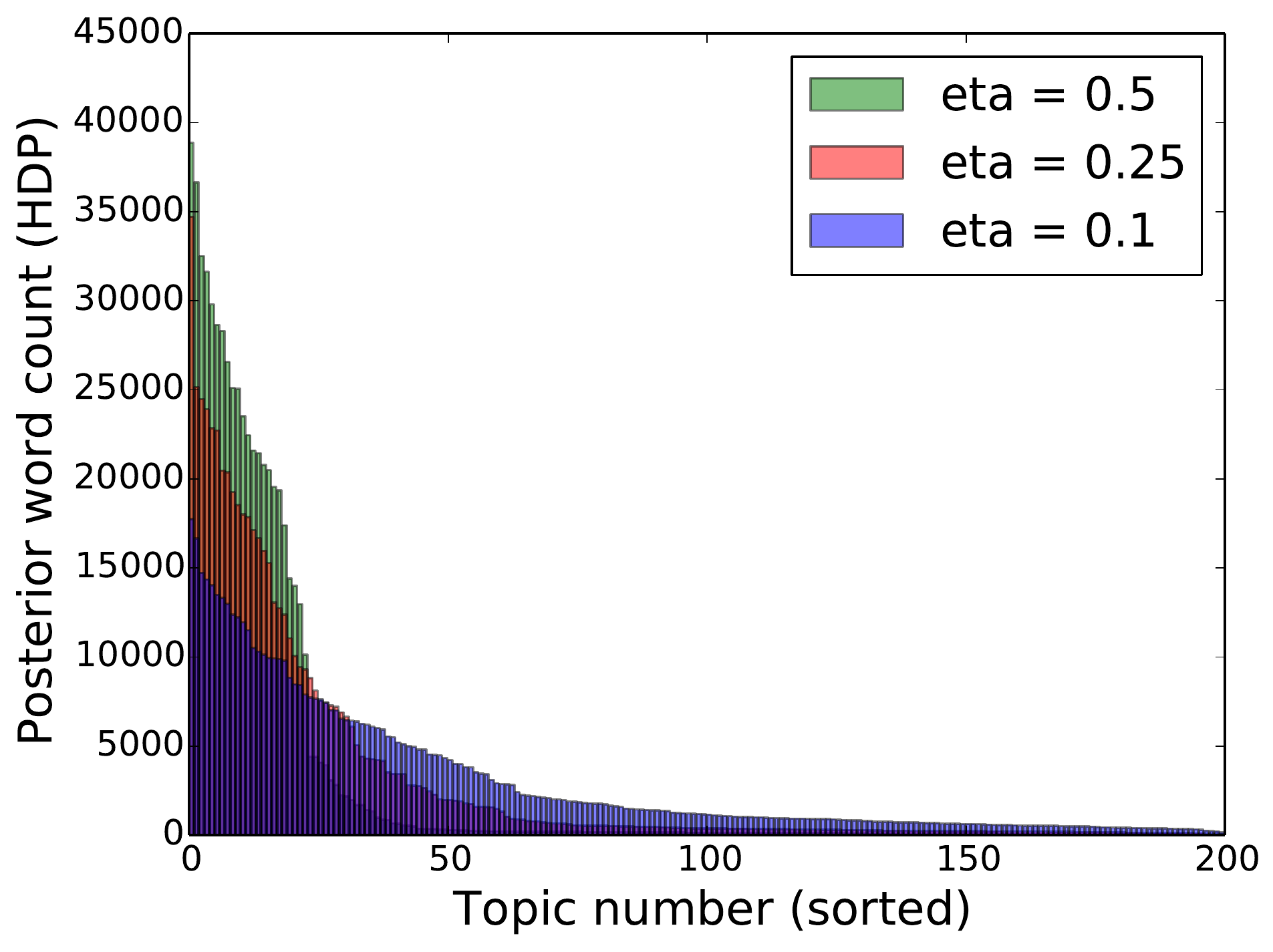}
	}
	\subfigure[HDSP]{
	\includegraphics[width=0.45\linewidth]{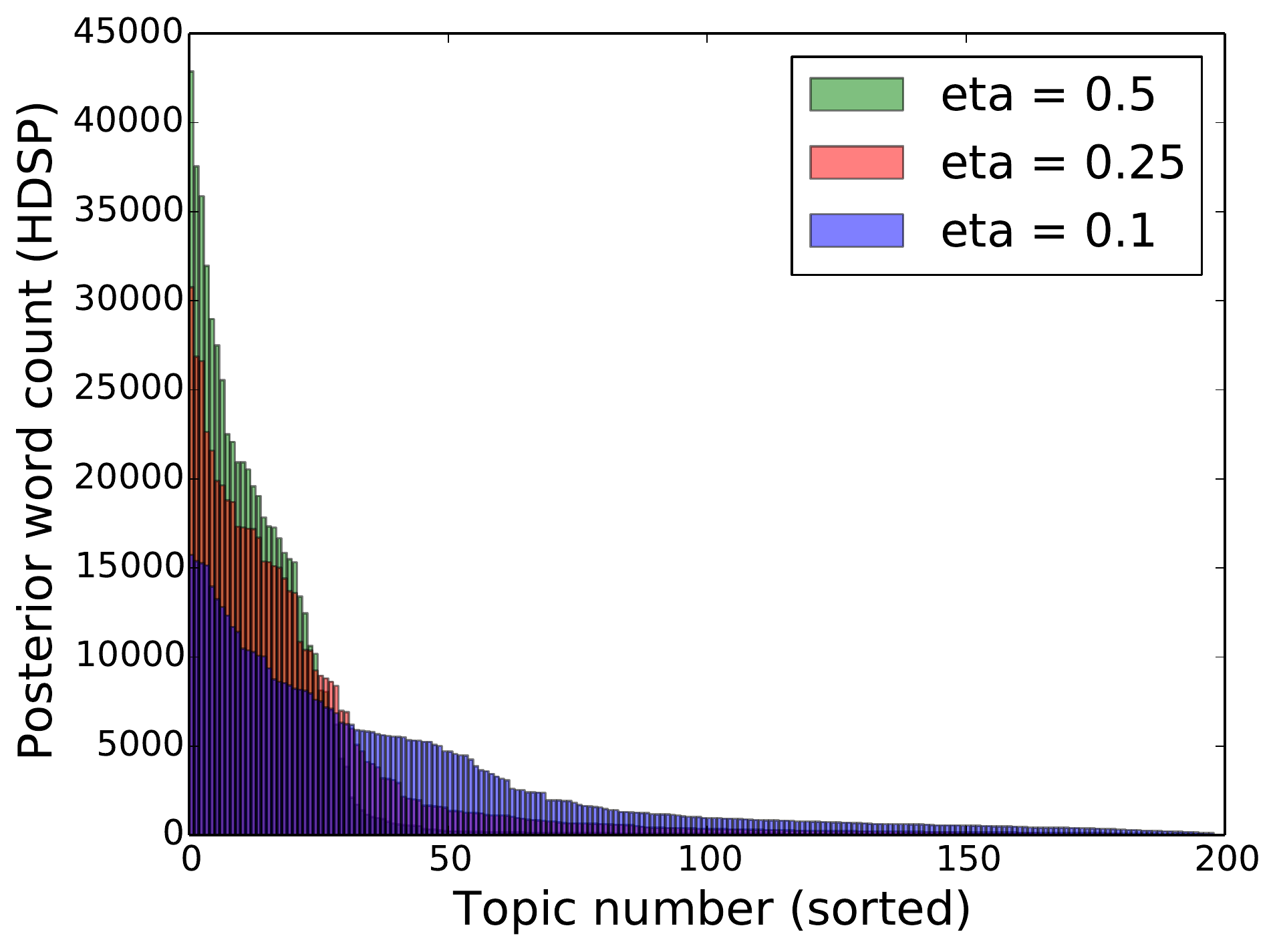}	
	}
	\caption{\label{fig:effective_topic} Posterior number of words. We set the truncation level to 200, but only a few topics are used during inference.}
\end{figure}

Like the HDP and other nonparametric topic models, our model also uses only a few topics even though we set the truncation level to 200. Figure \ref{fig:effective_topic} shows the posterior word count for the different values of the Dirichlet topic parameter $\eta$. As the result indicates our model uses 50 to 100 topics. The HDSP tends to use more topics than the HDP.

\vskip 0.2in
\bibliography{hdsp}

\end{document}